\documentclass[11pt]{article}

\usepackage[letterpaper,margin=1in]{geometry}

\usepackage{times}
\usepackage{helvet}
\usepackage{courier}

\usepackage[numbers]{natbib}

\usepackage[utf8]{inputenc}
\usepackage[T1]{fontenc}
\usepackage{hyperref}
\usepackage{url}
\usepackage{booktabs}
\usepackage{amsfonts}
\usepackage{amssymb}
\usepackage{amsmath}
\usepackage{amsthm}
\usepackage{nicefrac}
\usepackage{microtype}
\usepackage{xcolor}
\usepackage{enumitem}
\usepackage{bm}
\usepackage{graphicx}
\usepackage{subcaption}
\usepackage[normalem]{ulem}
\usepackage{tcolorbox}
\usepackage{wrapfig}
\usepackage{algorithm}
\usepackage{algorithmic}

\newcommand{\vphi}{{\boldsymbol{\phi}}}

\newcommand{\bR}{\mathbb{R}}
\newcommand{\bx}{\mathbf{x}}
\newcommand{\bz}{\mathbf{z}}
\newcommand{\bo}{\mathbf{o}}

\newcommand{\prob}{p}
\newcommand{\cN}{\mathcal{N}}
\newcommand{\cL}{\mathcal{L}}
\newcommand{\bv}{\mathbf{v}}
\newcommand{\bI}{\mathbf{I}}

\newcommand{\bepsilon}{\mathbf{\epsilon}}

\usepackage{color-edits}
\addauthor{ms}{blue}
\addauthor{cn}{green!70!black}
\addauthor{ab}{violet}

\newenvironment{squishenumerate}
    {\begin{list}{\arabic{enumi}.}{%
    \usecounter{enumi}%
    \setlength{\itemsep}{0pt}%
    \setlength{\parsep}{0pt}%
    \setlength{\topsep}{0pt}%
    \setlength{\parskip}{0pt}%
    \setlength{\labelwidth}{.5in}%
    \setlength{\labelsep}{0.05in}%
    \setlength{\leftmargin}{.2in}}}
{\end{list}}

\definecolor{predictcol}{RGB}{27, 94, 165}
\definecolor{updatecol}{RGB}{176, 78, 25}
\definecolor{resamplecol}{RGB}{40, 130, 70}

\title{\huge Generative Model Proposal based Particle Filtering for Data Assimilation}
\vspace{1em}

\author{%
  \textbf{Chandni Nagda}$^{1}$\thanks{Correspondence to: \texttt{cnagda2@illinois.edu}} \quad
  \textbf{Mayank Shrivastava}$^{1}$ \quad
  \textbf{Gudrun Thorkelsdottir}$^{1}$ \\
  \textbf{Gan Zhang}$^{1}$ \quad
  \textbf{Morteza Mardani}$^{2}$ \quad
  \textbf{Arindam Banerjee}$^{1}$ \\[4pt]
  $^{1}$University of Illinois at Urbana-Champaign \quad
  $^{2}$NVIDIA
}

\date{}

\begin{document}

\maketitle

\begin{abstract}
Data assimilation models state dynamics conditioned on sequential observations, and has wide-ranging scientific applications. In the filtering setting, the goal is to model the posterior over the current state given all observations so far. Classical solutions typically make simplifying distributional or functional assumptions, e.g., linear-Gaussian systems, which can be inaccurate in many scenarios. In principle, particle filters (PFs) remove these assumptions, yet often collapse in high dimensions.
Recent generative approaches learn conditional state transitions, but without principled Bayesian updates they do not recover the correct filtering posterior and can accumulate error over long horizons. In this work, we introduce Flow Proposal Particle Filters (FPPF), which learn a conditional generative model based proposal approximating the variance-minimizing optimal proposal for particle propagation. Conditioning on observations steers particles toward high-likelihood regions before weighting, reducing weight variance and delaying degeneracy. Since our proposal admits tractable likelihood evaluation, FPPF computes accurate importance weights and retains a Bayesian update step. We further extend FPPF to high-dimensional problems through localization strategies, adressing another standard PF failure mode. Extensive experiments on a variety of dynamical systems show that FPPF outperforms statistical baselines and other generative methods in non-linear, non-Gaussian, and high-dimensional regimes. Code is available at \url{https://github.com/cnagda/fppf/tree/main}.
\end{abstract}

\section{Introduction}
Data assimilation (DA) is the inverse problem of inferring the latent state trajectory $\bx_{1:T}$ of a dynamical system from imperfect observations $\bo_{1:T}$, using knowledge of the system's dynamics $p(\bx_t \mid \bx_{t-1})$ and an observation model $p(\bo_t \mid \bx_t)$. DA is fundamental to many scientific and engineering applications, including weather forecasting \citep{Lorenc1986AnalysisMethods, Rabier2000ECMWF4DVar, ASequentialEnsembleKalmanFilterforAtmosphericDataAssimilation}, climate analysis \citep{Dee2011ERAInterim, Hersbach2020ERA5}, agriculture \citep{DeWitVanDiepen2007CropEnKF, Dlamini15164066}, and motion tracking \citep{BarShalom2001TrackingNavigation, 772544}. In these domains, point estimates of latent states $\bx_t$ alone are inadequate. Since downstream decisions based on latent state estimates carry significant consequences, reliable uncertainty quantification requires recovering a full distribution over states. 

In this work, we focus on one of the most widely studied DA tasks called \emph{filtering} \cite{sarkka2013bayesian, chen2003bayesian}: at time $t$, given observations $\bo_{1:t}$, we seek the filtering posterior $p(\bx_t \mid \bo_{1:t})$. Bayesian filtering proceeds via a recursive prediction–update cycle, in which the state dynamics model defines the prior over the next state, and conditioning on the observation using Bayes' rule gives the posterior. When the dynamics and observation models are linear and the noise is Gaussian, this is solved exactly by the Kalman filter (KF) \citep{kalman1960kf}. To handle nonlinear dynamics, models like the Ensemble KF (EnKF) \citep{evensen1994enkf}, which are heavily used in operational settings \citep{HigherResolutioninanOperationalEnsembleKalmanFilter},  propagate an ensemble of samples through the true dynamics but retain Gaussian assumptions in the update step.

In principle, particle filters (PFs) \citep{gordon1993novel, doucet2000sequential} provide an appealing alternative, performing Bayesian filtering without linearity or Gaussianity assumptions by representing the posterior through a weighted ensemble of particles. However, they suffer from \emph{degeneracy}: in high-dimensional state spaces, the ensemble size required to prevent the weights from collapsing onto a few particles grows exponentially with the effective state dimension \citep{ObstaclestoHighDimensionalParticleFiltering, Bengtsson_2008}. Intuitively, particles propagated through the dynamics model often do not land in regions supported by the new observations, so most receive negligible weight. This curse of dimensionality has largely confined particle filters to low-dimensional problems.

In the general sequential importance resampling (SIR) framework~\citep{doucet2001smc},
one samples from an arbitrary proposal $q(\bx_t \mid \bx_{t-1}, \bo_t)$ and
corrects via importance weights. The variance-minimizing choice is the
\emph{optimal proposal} $q^\star(\bx_t \mid \bx_{t-1}, \bo_t) = p(\bx_t \mid
\bx_{t-1}, \bo_t)$, which steers particles toward regions supported by the new
observation before weighting \cite{doucet2000sequential}. However, $q^\star$ is hard to compute outside
special cases (such as linear-Gaussian models) which reintroduce restrictive
assumptions. A line of work has sought to move beyond this setting by
incorporating the incoming observation into how particles are selected or
propagated, including look-ahead particle selection \citep{pitt1999auxiliary},
implicit sampling via per-particle optimization \citep{chorin2009implicit,
morzfeld2012randommap}, and equal-weight constructions
\citep{vanleeuwen2010ewpf, ades2015ewpf}. However, these schemes either recover
$q^\star$ only under restrictive distributional structure or replace it with a
hand-designed surrogate, limiting their fidelity in the nonlinear, non-Gaussian
regimes that motivate PFs in the first place. What we would want instead is a
flexible approximation to $q^\star$ that keeps the full generality of PFs.

This is where recent ML-based approaches to DA become relevant. Recent methods
\citep{chen2025flowdas, huang2024diffda} learn the conditional $p(\bx_t \mid
\bx_{t-1}, \bo_t)$ directly with a generative model; note that this is precisely
the object the optimal proposal requires, free of Gaussian or linearity
assumptions. The prevailing strategy, however, is to apply the learned
conditional \emph{autoregressively}, rolling a state estimate forward by
repeatedly sampling from the generator. This amounts to propagating a single
unweighted particle: at no point is the filtering posterior $p(\bx_t \mid
\bo_{1:t})$ formed by combining the propagated prior $p(\bx_t \mid \bo_{1:t-1})$
with the likelihood $p(\bo_t \mid \bx_t)$ via Bayes' rule. Thus, each step
incurs a discrepancy with respect to the filtering posterior and empirically,
these purely generative rollouts rapidly accumulate error over long horizons.

In this work, we show that these two threads resolve each other. SIR filters
benefit from a flexible approximation of $q^\star$; generative DA methods learn
one, but deploy it naively. We propose Flow Proposal Particle Filters (FPPF),
which uses a learned conditional distribution as a proposal. Particles are
proposed in an observation-informed manner and then reweighted, so the filter
targets the true posterior even when the proposal is imperfect, while proximity
to $q^\star$ keeps weight variance low and delays degeneracy. While natural in
hindsight, realizing this step is nontrivial. We show how to learn $q_\phi(\bx_t \mid \bx_{t-1},
\bo_t)$ with conditional flow matching, which offers both efficient sampling and
exact likelihood evaluation via the instantaneous change-of-variables formula,
making the weight computation feasible. To extend the approach to
high-dimensional settings, where even well-proposed particles suffer weight
collapse, we introduce a localized variant, L-FPPF, whose velocity network
yields a per-site factorization of the proposal log-density and keeps both
computation and weighting local.

We evaluate FPPF and L-FPPF on chaotic dynamical systems,
including Lorenz-63, Lorenz-96, and the Kuramoto-Sivashinsky equation, under
both Gaussian and strongly non-Gaussian observation operators, against classical
filters, prior proposal learning methods, autoregressive generative DA, and
localized particle filters. FPPF consistently improves state estimation accuracy
and probabilistic calibration over these baselines, with the largest gains in
non-Gaussian regimes where Kalman updates fail. Unlike autoregressive rollouts,
it remains stable over long assimilation horizons. L-FPPF sustains these gains
as the state dimension grows, whereas global PFs degenerate.

We summarize our main contributions below:
\vspace{1em}
\begin{squishenumerate}
\item We propose FPPF, a filtering method that learns an observation-informed conditional generative model based proposal distribution approximating the variance-minimizing optimal proposal, and integrates it into the standard sequential importance resampling framework.
\item We instantiate the proposal with conditional flow matching, and show that likelihood evaluation can be leveraged to compute particle weights for a principled update step.
\item To extend the approach to high-dimensional settings, we propose L-FPPF, which uses a patch-based velocity network to obtain a factorization of the proposal log-density, extending localized particle filtering to general learned proposals.
\item We empirically demonstrate on chaotic dynamical systems (Lorenz-63, Lorenz-96, Kuramoto-Sivashinsky) that FPPF improves state estimation and probabilistic calibration over classical baselines and autoregressive generative models, and that L-FPPF scales to high-dimensional systems.
\end{squishenumerate}

The rest of the paper is organized as follows. Section 2 reviews related work, Section 3 provides background on Bayesian filtering and particle filters, Sections 4 and 5 present FPPF and its localized variant L-FPPF, Section 6 reports experiments, and we conclude in Section 7.

\section{Related Work}
\label{sec:related-work}

\paragraph{Classical data assimilation.}
The Kalman filter solves linear-Gaussian filtering exactly \citep{kalman1960kf}.
The EnKF and its transform variants propagate an ensemble through nonlinear
dynamics but retain a Gaussian update \citep{evensen1994enkf,
bishop2001adaptive, hunt2007efficient}, and are widely used in operational
weather prediction \citep{houtekamer2014higher, schraff2016kilometre}.
Variational methods such as 3D-Var and 4D-Var optimize a cost over the state trajectory
\citep{lorenc1986analysis, courtier1994strategy}, producing point estimates
rather than the posterior distributions we target.

Particle filters drop these assumptions entirely \citep{gordon1993novel}, but
the bootstrap proposal ignores the incoming observation, and the ensemble size
needed to prevent weight collapse grows exponentially with the effective state
dimension \citep{snyder2008obstacles, Bengtsson_2008}. The optimal proposal $q^\star$ minimizes incremental weight variance \citep{doucet2000sequential} but is hard to obtain. 
Classical mitigations approximate it indirectly: APF selects ancestors by observation
look-ahead without changing where particles land \citep{pitt1999auxiliary},
implicit sampling targets posterior modes via per-particle optimization
\citep{chorin2009implicit, morzfeld2012randommap}, and equal-weight constructions
hand-design moves toward the observation \citep{vanleeuwen2010ewpf, ades2015ewpf}. 
These recover $q^\star$ only under restrictive structure
or replace it with a surrogate. For spatially extended systems, localized PFs 
attack degeneracy along a different axis, computing weights and resampling 
per region from tapered local likelihoods \citep{poterjoy2016localized, penny2016local,
farchi2018comparison}. This construction is specific to the bootstrap PF,
whose weight is just the local observation likelihood; L-FPPF extends
localization to learned proposals.

\paragraph{Learned proposals for SMC.}
Early adaptive methods optimize proposals within parametric families with
closed-form densities \citep{cornebise2008adaptive, Branchini2020OptimizedAP}.
NASMC \citep{gu2015nasmc} and variational SMC \citep{naesseth2018variational,
le2018aesmc, maddison2017filtering} learn neural proposals by running a
particle filter inside the training loop, yielding gradient estimates that
suffer high variance when the proposal is poor or due to resampling. InfNN \citep{paige2016inference} trains offline by
maximum likelihood but restricts the proposal to a factorized autoregressive
form to keep its density tractable. In contrast, FPPF trains offline by
velocity regression on simulated tuples, imposes no distributional form, and
retains exact density evaluation via the instantaneous change-of-variables
formula.

\paragraph{Generative models for data assimilation.}
Score- and flow-based generative models have been applied to DA in two ways.
The first learns a conditional generator for the states themselves,
incorporating observations through inference-time methods such as guidance or inpainting: SDA samples short
trajectory windows given observations \citep{rozet2023scorebased}, while
DiffDA and FlowDAS apply a one-step conditional generator autoregressively
\citep{huang2024diffda, chen2025flowdas}. The guided iterates are not draws
from the filtering posterior, and in the autoregressive case this discrepancy
compounds, so errors accumulate over long rollouts
(Section~\ref{sec:experiments:ks}). A second line of work targets the analysis
distribution: EnSF \citep{bao_ensemble_2024} and EnFF \citep{transue_flow_2025} 
estimate the score or velocity of the filtering
density training-free and fold in observations through likelihood-gradient
guidance, latent variants improve
handling of sparse observations \citep{si2025latentensf, xiao2024ldensf}, and
the concurrent DAISI folds forecast information into a pretrained prior by
inverse sampling before guidance \citep{andrae2026daisi}. In both approaches
observations enter through guidance, an approximate correction to the
generative dynamics, with no mechanism to remove the resulting bias. Closest and concurrent
to our work, \citet{savary2026trainingfree} restore the Bayesian update for
this class, guiding a pretrained diffusion emulator with the likelihood score
inside a fully adapted auxiliary PF; FPPF learns a directly conditioned
proposal whose tractable density supplies the full SIR correction, keeping the
update valid even when the proposal is imperfect.

\paragraph{Learning the state-space model components.}
A complementary thread learns components of the state-space model itself.
\citet{brajard2020combining} emulate dynamics from EnKF analyses,
\citet{bocquet2020bayesian} jointly infer the dynamical model and the state,
and AD-EnKF \citep{chen2023autodifferentiable} differentiates through the EnKF
for joint state and parameter estimation. Differentiable particle filters
likewise learn transition, observation, and proposal end-to-end through
differentiable resampling \citep{karkus2018particle,
jonschkowski2018differentiable, corenflos2021differentiable, chen2023overview,
chen2024nfdpf}. These methods target model learning and parameter estimation;
we assume known dynamics and observation models and learn only the proposal,
which also avoids optimizing through the filter over long chaotic
trajectories.

\section{Background and Preliminaries}
\label{sec:prelims}

This section reviews the Bayesian filtering approach to data assimilation, and provides background on particle filters and the proposal distribution.

\subsection{Data Assimilation}
\label{sec:prelims:ssm}

We consider a dynamical system with latent state $\bx_t \in \bR^{d_x}$ and observations $\bo_t \in \bR^{d_o}$, where $t$ denotes the discrete time index. The evolution of the system is governed by a state transition model $\prob(\bx_t \mid \bx_{t-1})$. At each time step, observations are related to the latent state through the likelihood $p(\bo_t \vert \bx_t)$.
The goal is to estimate the \emph{filtering posterior} $\prob(\bx_t \mid \bo_{1:t})$, the distribution over the current state given all observations up to time $t$. In the Bayesian filtering framework~~\cite{jazwinski1970stochastic,anderson1979optimal,doucet2001smc,sarkka2013bayesian}, this distribution is computed recursively via a prediction-update cycle:
\begin{equation}
\label{eq:filtering}
\begin{aligned}
&\prob(\bx_t \mid \bo_{1:t-1}) = \int \prob(\bx_t \mid \bx_{t-1}) \prob(\bx_{t-1} \mid \bo_{1:t-1}) \, d\bx_{t-1}~, && \textcolor{predictcol}{\textbf{[\textsc{Predict}]}} \\
&\prob(\bx_t \mid \bo_{1:t}) \propto \prob(\bo_t \mid \bx_t) \prob(\bx_t \mid \bo_{1:t-1})~. && \textcolor{updatecol}{\textbf{[\textsc{Update}]}}
\end{aligned}
\end{equation}
For linear-Gaussian systems, i.e.\ when the transition dynamics are linear and the observation likelihood is Gaussian, the Kalman filter (KF)~\cite{kalman1960kf} provides an analytical solution to this recursion. Nonlinear extensions relax these assumptions approximately: the extended KF linearizes the dynamics and observation models~\cite{jazwinski1970stochastic}, the unscented KF propagates deterministically chosen sigma points~\cite{julier1997ukf}, and the ensemble KF~\cite{evensen1994enkf} replaces analytical covariance propagation with an ensemble, but retains a Gaussian update. In general, for nonlinear dynamics and non-Gaussian observation models, the prediction integral remains intractable and the update step admits no closed form, necessitating other inference methods.

\subsection{Particle Filters}
\label{sec:prelims:pf}
Particle filters (PFs) approximate the filtering posterior $\prob(\bx_t \mid \bo_{1:t})$ using a weighted ensemble of $N$ samples $\{ \bx_t^{(i)}, w_t^{(i)} \}_{i=1}^N$ with $\sum_i w_t^{(i)} = 1$, representing the posterior through the empirical measure $\sum_i w_t^{(i)} \delta_{\bx_t^{(i)}}$.

\paragraph{Bootstrap particle filter.}
The bootstrap particle filter (BPF)~\citep{gordon1993novel} instantiates the recursion of Eq.~\eqref{eq:filtering} via Monte Carlo. Given the weighted ensemble $\{\bx_{t-1}^{(i)}, w_{t-1}^{(i)}\}$ approximating $\prob(\bx_{t-1} \mid \bo_{1:t-1})$, the \textcolor{predictcol}{\textsc{Predict}} step propagates each particle through the transition dynamics, and the \textcolor{updatecol}{\textsc{Update}} step reweights by the new observation likelihood:
\begin{equation}
\label{eq:bpf}
\begin{aligned}
\bx_t^{(i)} &\sim \prob(\bx_t \mid \bx_{t-1}^{(i)})~, && \textcolor{predictcol}{\textbf{[\textsc{Predict}]}} \\
w_t^{(i)} &\propto w_{t-1}^{(i)} \, \prob(\bo_t \mid \bx_t^{(i)})~, \qquad \textstyle\sum_i w_t^{(i)} = 1~. && \textcolor{updatecol}{\textbf{[\textsc{Update}]}}
\end{aligned}
\end{equation}
Sampling from the transition realizes the prediction integral in Eq.~\eqref{eq:filtering}, and multiplying by the likelihood realizes the Bayesian update; the weight recursion is exactly importance sampling against the one-step prior. The BPF is appealing because it requires only the ability to simulate the dynamics and evaluate the likelihood, with no Gaussian or linearity assumptions.

\paragraph{Weight degeneracy.}\label{sec:prelims:degeneracy} The standard BPF suffers from \emph{degeneracy} in high-dimensional spaces because particles propagated through the transition prior often fail to land in regions supported by the new observations. Consequently, most particles receive negligible weight, causing the mass to collapse onto a single particle during the \textcolor{updatecol}{\textsc{Update}} step. The severity of this collapse is captured by the variance of the log-likelihoods across the propagated ensemble, \(\tau^2 := \mathrm{Var}\big[\log \prob(\bo_t \mid \bx_t^{(i)})\big]\): \citet{Bengtsson_2008} and \citet{ObstaclestoHighDimensionalParticleFiltering} show that the largest normalized weight tends to \(1\) unless the ensemble size satisfies \(\log N \gtrsim \tau^2/2\). Since \(\tau^2\) grows roughly linearly with the effective dimension of the system, avoiding degeneracy demands \(N \gtrsim \exp(\tau^2/2)\) particles, a requirement that quickly becomes intractable and fundamentally limits the use of BPF in high-dimensional domains \citep{snyder2015performance}.

\paragraph{Resampling.} To mitigate weight degeneracy, which occurs even in low-dimensional problems, the BPF \emph{resamples}. This step draws $N$ particles with replacement proportional to $w_t^{(i)}$, duplicating high-weight particles, discarding low-weight ones, and resetting all weights to $1/N$ \citep{gordon1993novel, liu1998sequential}. Stochastic model dynamics can subsequently restore particle diversity. However, if the weights collapse completely before resampling, the algorithm duplicates a single surviving particle $N$ times, an unrecoverable loss of diversity known as \emph{sample impoverishment} \cite{arulampalam2002tutorial}. Therefore, keeping weights balanced is a fundamental problem, and designing better proposal distributions provides a path to achieving this.

\paragraph{Proposal distribution.} The sequential importance resampling (SIR) framework \citep{doucet2000sequential, doucet2009tutorial} generalizes BPF by drawing particles from a proposal $q(\bx_t \mid \bx_{t-1}, \bo_t)$ that may depend on the current observation, with weights updated as
\begin{equation}
\label{eq:sir_weight}
w_t^{(i)} \propto w_{t-1}^{(i)} \, \frac{\prob(\bo_t \mid \bx_t^{(i)}) \, \prob(\bx_t^{(i)} \mid \bx_{t-1}^{(i)})}{q(\bx_t^{(i)} \mid \bx_{t-1}^{(i)}, \bo_t)} \, .
\end{equation}
BPF corresponds to the choice $q = \prob(\bx_t \mid \bx_{t-1})$, which ignores $\bo_t$ at the \textcolor{predictcol}{\textsc{Predict}} stage; the incremental weight then reduces to the likelihood $\prob(\bo_t \mid \bx_t^{(i)})$ (Eq. \eqref{eq:bpf}).
A natural way to do better is to fold the current observation into the proposal itself, drawing from $\prob(\bx_t \mid \bx_{t-1}, \bo_t)$. Expanding this density by Bayes' rule, and using that $\bo_t$ depends on the state only through $\bx_t$ so that $\prob(\bo_t \mid \bx_t, \bx_{t-1}) = \prob(\bo_t \mid \bx_t)$, gives
\begin{equation}
\label{eq:optimal_proposal}
q^\star(\bx_t \mid \bx_{t-1}, \bo_t) = \prob(\bx_t \mid \bx_{t-1}, \bo_t)
= \frac{\prob(\bo_t \mid \bx_t, \bx_{t-1}) \, \prob(\bx_t \mid \bx_{t-1})}{\prob(\bo_t \mid \bx_{t-1})}
= \frac{\prob(\bo_t \mid \bx_t) \, \prob(\bx_t \mid \bx_{t-1})}{\prob(\bo_t \mid \bx_{t-1})} \, .
\end{equation}
Substituting $q^\star$ into the weight update collapses the numerator against the proposal, leaving
\begin{equation}
\label{eq:opt-weight}
w_t^{(i)} \propto w_{t-1}^{(i)} \, \prob(\bo_t \mid \bx_{t-1}^{(i)}) \, ,
\end{equation}
so that the weight at time $t$ depends only on the ancestor particle $\bx_{t-1}^{(i)}$ and not on the sampled $\bx_t^{(i)}$.
In this sense, the proposal $q^\star$ is often termed ``optimal'' \citep{doucet2000sequential}: conditional on $\bx_{t-1}^{(i)}$ and $\bo_t$, the incremental weight is independent of the random draw $\bx_t^{(i)}$, and hence has zero variance with respect to the proposal. This optimality is therefore local; it does not imply that the resulting particle filter performs optimally overall, but only that it minimizes the variance $\mathrm{Var}\!\big(w_t^{(i)}/w_{t-1}^{(i)} \mid \bx_{t-1}^{(i)}, \bo_t\big)$ introduced at the current sampling step.
Intuitively, this proposal shifts the effect of the likelihood from the weighting step into the sampling step, so that particles are drawn preferentially in regions of high posterior probability, which reduces the weight degeneracy problem, although it does not completely solve it.

\paragraph{High-dimensional particle filtering.}
The optimal proposal above admits a tractable closed form only in special cases, most notably linear--Gaussian state space models, where $q^\star$ is Gaussian and reproduces the Kalman analysis~\citep{doucet2000sequential}. Outside these restricted settings, evaluating the normalizing integral in $\prob(\bo_t \mid \bx_{t-1})$ is intractable, and even if one can sample exactly from $q^\star$, the ensemble size required to avoid weight collapse still grows too rapidly for many high-dimensional problems~\citep{ObstaclestoHighDimensionalParticleFiltering, snyder2015performance, Bengtsson_2008}.

A wide range of approaches has been proposed to mitigate this degeneracy~\citep{vanleeuwen2019review}. One primary direction builds more informative proposals that fold the observation into propagation, but approximates $q^\star$ only indirectly. The auxiliary particle filter (APF), for instance, performs a one-step look-ahead through a first-stage weight
\begin{equation}
\lambda_t^{(i)} \propto w_{t-1}^{(i)} \, \prob\big(\bo_t \mid \boldsymbol{\mu}_t^{(i)}\big), \qquad \boldsymbol{\mu}_t^{(i)} = \mathbb{E}\big[\bx_t \mid \bx_{t-1}^{(i)}\big],
\end{equation}
that ranks ancestors by the observation likelihood at their predicted mean without changing where particles are drawn. Other approaches such as implicit sampling and equal-weight constructions (Section~\ref{sec:related-work}), also recover $q^\star$ only under restrictive distributional structure.

A second primary direction exploits spatial locality. Localized particle filters partition the domain into regions $\{\mathcal{B}_b\}$ with centers $c_b$ and weight each from a tapered local likelihood,
\begin{equation}
\log w_{b,t}^{(i)} \propto \log w_{t-1}^{(i)} + \sum_{o} G\!\left(\frac{d(o, c_b)}{r_{\mathrm{loc}}}\right) \log \prob\big(\bo_{t,o} \mid \bx_t^{(i)}\big),
\end{equation}
so that a region's effective dimension is governed by the radius $r_{\mathrm{loc}}$ rather than $d_x$~\citep{poterjoy2016localized, rebeschini2015local}. This localization, however, is formulated for the bootstrap proposal, whose weight is simply the observation likelihood. 

Our method builds on both threads, learning an approximation of $q^\star$ with a tractable weight correction and extending it to a localized setting for high-dimensional problems.

\section{Flow Proposal Particle Filter (FPPF)}
\label{sec:method}

We now present Flow Proposal Particle Filter (FPPF), our approach for learning proposal distributions that approximate the locally optimal proposal. We parameterize the proposal using conditional flow matching (FM)~\citep{lipman2023flow, liu2023flow}, a generative modeling framework that learns a time-dependent velocity field to transport samples from a simple base distribution to a target distribution via an ordinary differential equation (ODE). FM admits stable and efficient training via simple regression and tractable exact density evaluation for importance weight computation. The full method is summarized in Algorithm \ref{alg:FPPF}.

\begin{algorithm}[t]
\caption{Flow Proposal Particle Filter (FPPF)}
\label{alg:FPPF}
\begin{algorithmic}[1]
\REQUIRE Trained velocity field $v_\vphi$, observations $\bo_{1:T}$,
particles $N$, threshold $N_{\mathrm{eff}}$
\STATE Initialize: $\bx_0^{(i)} \sim p_0(\bx)$, $w_0^{(i)} = 1/N$ for $i = 1, \ldots, N$
\FOR{$t = 1, \ldots, T$}
    \FOR{$i = 1, \ldots, N$}
        \STATE \textcolor{predictcol}{\textbf{[\textsc{Predict}]}} Sample $\bz^{(i)}(0) \sim \cN(\mathbf{0}, \bI)$;
        integrate Eq.~\eqref{eq:proposal_ode} to obtain $\bx_t^{(i)}$
        \STATE \textcolor{predictcol}{\textbf{[\textsc{Predict}]}} Evaluate
        $\log q_\vphi(\bx_t^{(i)} \mid \bx_{t-1}^{(i)}, \bo_t)$
        \STATE \textcolor{updatecol}{\textbf{[\textsc{Update}]}}
        $\tilde{w}_t^{(i)} \gets w_{t-1}^{(i)} \cdot \prob(\bo_t \mid \bx_t^{(i)}) \cdot \prob(\bx_t^{(i)} \mid \bx_{t-1}^{(i)}) \,/\, q_\vphi(\bx_t^{(i)} \mid \bx_{t-1}^{(i)}, \bo_t)$
    \ENDFOR
    \STATE Normalize: $w_t^{(i)} \gets \tilde{w}_t^{(i)} / \sum_{j=1}^N \tilde{w}_t^{(j)}$
    \STATE Compute $\mathrm{ESS}_t \gets 1 / \sum_{i=1}^N (w_t^{(i)})^2$
    \IF{$\mathrm{ESS}_t < N_{\mathrm{eff}}$}
        \STATE \textcolor{resamplecol}{\textbf{[\textsc{Resample}]}} Draw
        $\{\bx_t^{(i)}\}$ with replacement according to $\{w_t^{(i)}\}$;
        reset $w_t^{(i)} \gets 1/N$
    \ENDIF
\ENDFOR
\STATE Return weighted particles $\{(\bx_t^{(i)}, w_t^{(i)})\}_{t=1}^T$
\end{algorithmic}
\end{algorithm}

\subsection{Learning the Proposal Distribution}
\label{sec:method:learning}
We learn a conditional proposal distribution for the filtering problem of the form
\(
q_\phi(\bx_t \mid \bx_{t-1}, \bo_t)
\),
used to propagate particles from time $t-1$ to time $t$ given the current observation.
It is a flexible approximation to the locally optimal proposal
$q^\star(\bx_t \mid \bx_{t-1}, \bo_t)$ of Eq.~\eqref{eq:optimal_proposal}.
The proposal is parameterized implicitly via a conditional velocity field
$v_\phi(\bz(s), s; \bx_{t-1}, \bo_t)$
that defines a deterministic flow transporting samples from a simple base distribution to the proposal distribution.

Here $t$ is the discrete filtering time indexing the state sequence, while
$s \in [0,1]$ is an auxiliary flow time internal to the generative model; one
full integration in $s$ produces a single proposal draw at filtering step $t$.
We define $\bz(s)$ by the ordinary differential equation, holding the
conditioning variables $(\bx_{t-1}, \bo_t)$ fixed:
\begin{equation}
    \frac{d\bz(s)}{ds}
    =
    v_\phi\!\left(\bz(s), s; \bx_{t-1}, \bo_t \right),
    \qquad
    \bz(0)  \sim \mathcal{N}(\mathbf{0}, \bI).
    \label{eq:proposal_ode}
\end{equation}
Integrating~\eqref{eq:proposal_ode} from $s=0$ to $s=1$ yields a proposal sample
\(
\bx_t := \bz(1)
\).

\paragraph{Network conditioning.}
We parameterize $v_\phi$ as a neural network taking as input both the previous
state $\bx_{t-1}$ and the current observation $\bo_t$ to approximate the optimal
proposal. We use simple concatenation, providing $[\bx_{t-1}; \bo_t]$ as additional input alongside the noised state $\bx_s$ and flow time $s$. Alternative conditioning mechanisms such as 
FiLM \citep{Perez_Strub_deVries_Dumoulin_Courville_2018} or cross-attention can also be employed. 
We choose direct conditioning, rather than guidance-based approaches
\citep{song2021scorebased, chen2025flowdas} that incorporate observations at
inference time by adding an approximate likelihood-gradient term to the
generative drift. Inference-time guidance is appealing because the same generative prior serves
any observation model without retraining; however, in practice the guidance
scale is sensitive to the observation operator and noise level and must be
retuned per setting
(too small and the observation is under-enforced, too large and the
approximate gradient distorts the samples) \citep{ho2021classifierfree, saharia2022photorealistic}. It also keeps density evaluation simple: with no auxiliary likelihood-gradient
term in the dynamics, the proposal density follows from applying the
change-of-variables formula to $v_\phi$ alone (Section~\ref{sec:method:weights}).

To prevent the model from over-relying on observations, which could degrade
performance when they are noisy or uninformative, we apply observation
dropout during training (similar to ~\cite{ho2021classifierfree}): for each training tuple
independently, with probability $p_{\mathrm{cond}}$ we replace the entire
observation vector $\bo_t$ with zeros, while leaving $\bx_{t-1}$ intact. This
exposes the network to both observation-conditioned and observation-free inputs,
encouraging it to also leverage dynamical information from $\bx_{t-1}$ rather
than relying solely on $\bo_t$. We use $p_{\mathrm{cond}} = 0.1$.
Similarly, to prevent over-reliance on the previous state $\bx_{t-1}$, we
corrupt it during training with a coordinate-masking schedule. 
At training step $k$ of $K$ total, with probability
$p_{\mathrm{mask}}(k) = \max\!\big(p_{\min},\, p_0(1 - k/K)\big)$ we zero a
uniformly random subset of $\lceil r_{\mathrm{mask}}\, d_x \rceil$ coordinates
of $\bx_{t-1}$ before passing it to the network, and otherwise use it intact.
The masking probability is annealed linearly from $p_0$ to a floor $p_{\min}$
over training, so the network is forced to recover masked coordinates from
$\bo_t$ early on, when it would otherwise lock onto the highly predictable
dynamics, and is relaxed toward the clean state as training converges. All masking
is disabled at inference.

\paragraph{Training data.}
For a given dynamical system, using a transition model $p(\bx_t \mid \bx_{t-1})$ and an observation model $p(\bo_t \mid \bx_t)$, we simulate trajectories and form supervised training tuples
$(\bx_{t-1}, \bo_t, \bx_t)$, where $\bo_t$ is sampled from $p(\bo_t \mid \bx_t)$. Specific choices for trajectory generation are in Section \ref{sec:experiments}.

\paragraph{Conditional flow matching objective.}
Following \citet{liu2023flow}, we adopt linear interpolation paths between
paired base and target samples, which induce a target velocity
$\bz(1) - \bz(0)$ that $v_\vphi$ regresses onto. For each training tuple, we sample noise $\bz(0) \sim \cN(\mathbf{0}, \bI)$, set $\bz(1) := \bx_t$, draw $s \sim \mathrm{Unif}[0,1]$, and form the linear interpolant $\bz(s) = (1-s)\bz(0) + s\bz(1)$, whose constant velocity is $\bz(1) - \bz(0)$. 
We train $v_\vphi$ to regress onto this target by minimizing
\begin{equation}
\cL_{\mathrm{FM}}(\vphi) =
\mathbb{E}_{(\bx_{t-1}, \bo_t, \bx_t),\, \bz(0),\, s}
\left[
\big\| v_\vphi\big(\bz(s), s;\, \bx_{t-1}, \bo_t\big) - \big(\bz(1) - \bz(0)\big) \big\|_2^2
\right],
\label{eq:fm_objective}
\end{equation}
where $(\bx_{t-1}, \bo_t, \bx_t)$ is a training tuple from the simulated
trajectories (with $\bz(1) := \bx_t$), $\bz(0) \sim \cN(\mathbf{0}, \bI)$,
$s \sim \mathrm{Unif}[0,1]$, and $\bz(s) = (1-s)\bz(0) + s\,\bz(1)$.

\subsection{Importance Weight Computation}
\label{sec:method:weights}
Given particles $\{(\bx_{t-1}^{(i)}, w_{t-1}^{(i)})\}_{i=1}^N$ approximating
$\prob(\bx_{t-1} \mid \bo_{1:t-1})$ (weights initialized to $1/N$), we sample
$\bx_t^{(i)} \sim q_\vphi(\bx_t \mid \bx_{t-1}^{(i)}, \bo_t)$
via~\eqref{eq:proposal_ode}. We then apply the SIR update of
Eq.~\eqref{eq:sir_weight} with our learned proposal $q_\vphi$
(Algorithm~\ref{alg:FPPF}, line~7):
\begin{equation}
    \tilde{w}_t^{(i)} = w_{t-1}^{(i)} \frac{\prob(\bo_t \mid \bx_t^{(i)}) \, \prob(\bx_t^{(i)} \mid \bx_{t-1}^{(i)})}{q_\vphi(\bx_t^{(i)} \mid \bx_{t-1}^{(i)}, \bo_t)}.
    \label{eq:weight_update}
\end{equation}
Since the model dynamics and observation model are known, the numerator is straightforward to compute. A key property of FM is exact density evaluation via the instantaneous change of variables formula~\citep{lipman2024flowmatchingguidecode}. For a sample $\bx_t$ generated by integrating Eq.~\eqref{eq:proposal_ode} from initial noise $\bz(0)$,
\begin{equation}
    \log q_\vphi(\bx_t \mid \bx_{t-1}, \bo_t) = \log p_0(\bz(0)) - \int_0^1 \mathrm{Tr}\left(\frac{\partial v_\vphi}{\partial \bz(s)}\right) ds,
\end{equation}
where $p_0 = \cN(\mathbf{0}, \bI)$ is the base distribution. Computing the exact trace requires $d_x$ backward passes through the network, which becomes prohibitive in high dimensions. We instead use the Hutchinson trace estimator, as in previous works \citep{lipman2024flowmatchingguidecode}:
$\mathrm{Tr}(\mathbf{J}) = \mathbb{E}_{\boldsymbol{\epsilon}}\left[\boldsymbol{\epsilon}^\top \mathbf{J} \boldsymbol{\epsilon}\right],$
where $\boldsymbol{\epsilon}$ satisfies $\mathbb{E}[\boldsymbol{\epsilon}\bepsilon^\top] = \bI$. The vector-Jacobian product $\boldsymbol{\epsilon}^\top \mathbf{J}$ is computed efficiently via a single backward pass. We use Rademacher random vectors ($\epsilon_j \in \{-1, +1\}$ uniformly). For a
fixed matrix $\mathbf{J}$, the single-probe estimator variance is
$2\|\mathbf{J}\|_F^2$ for Gaussian probes and
$2\sum_{i \neq j} J_{ij}^2$ for Rademacher probes
\citep{hutchinson1990stochastic}; the latter removes the diagonal contribution
and is never larger, which is why Rademacher probes are standard and match our
empirical observation of lower variance. 
Finally, resampling is triggered when the effective sample size
$\mathrm{ESS} = 1 / \sum_i (w_t^{(i)})^2$ falls below a threshold
$N_{\mathrm{eff}}$ \citep{liu1998sequential}.

\paragraph{Temporally sparse observations.}
When no observation is available at a given step, we default to the bootstrap
transition, sampling $\bx_t^{(i)} \sim \prob(\bx_t \mid \bx_{t-1}^{(i)})$ and
omitting the observation update, and apply the learned proposal and SIR
correction only at steps where an observation arrives.

\section{Localized Flow Proposal Particle Filter (L-FPPF)}
\label{sec:method2}

Particle filters require ensembles that grow exponentially with the effective
state dimension (Section~\ref{sec:prelims:pf}), even under the optimal proposal
\citep{snyder2008obstacles, snyder2015performance}. One solution is
\emph{localization}, which mitigates this
by exploiting sparse conditional dependence. Such structure is naturally described by a graphical
model, i.e. the filtering posterior factorizes according to a graph whose nodes are
the state coordinates (which we call \emph{sites}) and whose edges encode direct
statistical dependence, so that each site is conditionally independent of the
rest given its neighbors \citep{lauritzen1996graphical}. Sparse graphs of this kind are common
in high-dimensional dynamical systems and can be learned from data when not known
a priori \citep{drton2017structure, runge2019inferring}.  In the spatially extended systems
we study the graph is both known and local, since each site couples directly only
to its spatial neighbors.

\paragraph{Localized Particle Filters (LPFs).} A class of methods known as localized particle filters (LPFs) exploit this structure to 
avoid collapse
\citep{poterjoy2016localized, penny2016local, farchi2018comparison}. Exact
inference over a graphical model is tractable only when the graph has small
treewidth, a condition the dense couplings of a filtering posterior generally
violate, so localized PFs approximate the single global update with many
overlapping local ones. Each local update is restricted by a localization radius,
which bounds its effective dimension independently of $d_x$. Because the required ensemble size scales only with this 
local effective dimension, localization prevents weight collapse
and drastically reduces the necessary number of particles for a high-dimensional problem. This whole
construction rests on the bootstrap proposal. Because the bootstrap importance
weight is just the observation likelihood $\prob(\bo_t \mid \bx_t^{(i)})$
(Eq.~\eqref{eq:bpf}), it factorizes over sites whenever the observation noise is
independent across coordinates, and the per-site terms can then be tapered by
spatial distance and aggregated block by block. 

A learned proposal breaks this decomposition. 
We avoid the bootstrap proposal in
the first place because it ignores the incoming observation and degenerates quickly in
high dimensions, but moving to a learned $q_\phi$ creates a different obstacle. A
global velocity network couples all sites at once, so the proposal density
$\log q_\phi$ that appears in the denominator of the importance weight
(Eq.~\eqref{eq:sir_weight}) no longer factorizes, and without that factorization
the weights cannot be localized. We therefore localize the proposal itself,
restricting the network so that $\log q_\phi$ factorizes across sites by
construction.

\paragraph{Localized velocity field.}
We replace the global velocity network with a patch network \(u_\phi\) shared across all state dimensions, which we call \emph{sites}. For the spatially extended systems considered here, the sites lie on a periodic one-dimensional lattice, so neighborhoods wrap around at the boundary. Thus, the radius-\(r\) neighborhood of site \(j\) is \(W_j^r=\{j-r,\ldots,j+r\}\), where indices outside \(\{1,\ldots,d_x\}\) are wrapped periodically. Writing \(c_t=(\bx_{t-1},\bo_t)\), the localized velocity is
\begin{equation}
v_{\phi,j}(\bz(s),s;c_t)
=
u_\phi\!\left(\bz_{W_j^r}(s),\bx_{t-1,W_j^r},\bo_{t,W_j^r},s\right).
\label{eq:local_velocity}
\end{equation}

This architecture is motivated by localized generative modeling. For distributions with sparse or approximately sparse graphical dependence, \citet{gottwald2025localized} show that score components can be approximated from local graph neighborhoods, with localization error decaying as the neighborhood radius grows. Their analysis applies to score-based diffusion models, while our proposal is trained by conditional flow matching along rectified-flow interpolation paths; we therefore use this result as motivation for imposing locality on the proposal velocity, leaving a formal analysis of localized rectified-flow proposals to future work. In the spatial systems considered here, the transition and observation mechanisms are local or coordinate-wise, so the observation-informed proposal \(q^\star(\bx_t\mid\bx_{t-1},\bo_t)\) is expected to be dominated by local dependencies.

The locality constraint also makes proposal-density evaluation tractable. Since \(v_{\phi,j}\) depends only on \(\bz_{W_j^r}\), the velocity Jacobian is sparse: \(\partial v_{\phi,j}/\partial z_k=0\) whenever \(k\notin W_j^r\). Applying the instantaneous change-of-variables formula used for the global FPPF proposal, the localized proposal log-density decomposes as
\begin{equation}
\log q_\phi(\bx_t\mid\bx_{t-1},\bo_t)
=
\sum_{j=1}^{d_x}\ell_j,
\qquad
\ell_j
=
\log \mathcal{N}(z_j(0);0,1)
-
\int_0^1
\frac{\partial v_{\phi,j}}{\partial z_j}(\bz(s),s;c_t)\,ds .
\label{eq:logq_local_trace}
\end{equation}
This follows because the base distribution factorizes across coordinates and the change-of-variables trace is the sum of diagonal Jacobian entries. The site-wise terms \(\ell_j\) provide the proposal-density contribution used by the localized SIR correction: in the block-wise update, each block \(B_b\) uses \(\sum_{j\in B_b}\ell_j\), together with the corresponding local transition and tapered observation-likelihood terms.

In implementation, local patches for all sites are formed as a single batch, and \(u_\phi\) is applied once to produce all site velocities in parallel. The diagonal terms \(\partial v_{\phi,j}/\partial z_j\) are computed exactly by differentiating each scalar site output with respect to the center coordinate of its input patch. Thus, L-FPPF replaces the Hutchinson trace estimator used by global FPPF with an exact deterministic trace computation whose cost scales with the local patch size. Training and implementation details are provided in Appendix~\ref{app:architectures}.

\paragraph{Block-wise update step.}
Following the standard LPF recipe \citep{poterjoy2016localized, farchi2018comparison},
we partition the state sites into $B$ disjoint contiguous blocks
$\{\mathcal{B}_b\}_{b=1}^B$ with centers $c_b$. The blocks determine where resampling
decisions are made, and are distinct from the observation neighborhoods used to form the
weights. Each block's weight uses observations within a smooth distance-based taper around
$c_b$, so the neighborhoods of nearby blocks overlap even though the blocks themselves do
not.

We obtain the block weights by localizing the global SIR log-weight (Eq.~\eqref{eq:weight_update}).
We assume that the observation likelihood and the transition density to factorize over sites:
\begin{equation}
\log \prob(\bo_t \mid \bx_t) = \sum_{m=1}^{d_o} \log \prob(o_{t,m} \mid \bx_t),
\qquad
\log \prob(\bx_t \mid \bx_{t-1}) = \sum_{j=1}^{d_x} \log \prob(x_{t,j} \mid \bx_{t-1}).
\label{eq:factorization}
\end{equation} 
This further requires that the observation and process noise are
independent across coordinates (diagonal Gaussian noise in our experiments). Together with
the per-site proposal terms
$\ell_j$ from Eq.~\eqref{eq:logq_local_trace}, every ingredient of the SIR correction is then
site-wise, so it can be localized by tapering the observation likelihoods and restricting
the transition/proposal correction to each block. Writing
$\rho_{b,m} = G\!\left(d_{m,b} / r_{\mathrm{loc}}\right)$ for the taper between observation
$m$ and block center $c_b$, the \textcolor{updatecol}{\textsc{Update}} step becomes
\begin{equation}
\log \tilde{w}_{b,t}^{(i)}
= \log w_{t-1}^{(i)}
+ \sum_{m=1}^{d_o} \rho_{b,m} \log \prob(o_{t,m} \mid \bx_t^{(i)})
+ \sum_{j \in \mathcal{B}_b}\!\left[\log \prob(x_{t,j}^{(i)} \mid \bx_{t-1}^{(i)}) - \ell_j^{(i)}\right],
\label{eq:lfppf_weights}
\end{equation}
where $G$ is the Gaspari--Cohn taper \citep{gaspari1999construction}, $r_{\mathrm{loc}}$ is
the localization radius, and $d_{m,b}$ is the distance between observation $m$ and block
center $c_b$ on the spatial dependency graph (the wrapped lattice distance in our
experiments). The first sum is the tapered observation log-likelihood used by the
localized bootstrap PF; the second is the block-restricted SIR correction, made tractable
by Eq.~\eqref{eq:logq_local_trace} and absent from prior LPFs, which all use the bootstrap
transition as proposal. This is a localized approximation to the global SIR update, not an
exact factorization of the filtering posterior across blocks. We also note that the
localized velocity restricts how each site depends on its neighbors, but the flow ODE is
integrated jointly over all coordinates.

For each block $b$, the unnormalized weights in Eq.~\eqref{eq:lfppf_weights} are normalized
over particles, $\bar{w}_{b,t}^{(i)} = \tilde{w}_{b,t}^{(i)} / \sum_{n=1}^N \tilde{w}_{b,t}^{(n)}$.
We then resample independently within each block: let $a_b^{(i)}$ denote the ancestor index
selected for output particle $i$ from the normalized block weights $\{\bar{w}_{b,t}^{(n)}\}_{n=1}^N$.
Following \citet{farchi2018comparison}, the selected ancestor indices are reordered to
preserve particle identity where possible. A new full-state particle is then assembled
site by site,
\begin{equation}
x_{t,j}^{(i),\,\mathrm{new}} = x_{t,j}^{(a_{b(j)}^{(i)})},
\label{eq:reconstruction}
\end{equation}
where $b(j)$ is the block containing site $j$. This replaces the ESS-triggered global
\textcolor{resamplecol}{\textsc{Resample}} of Algorithm~\ref{alg:FPPF} with block-wise
resampling at every observation time.

\section{Experiments}
\label{sec:experiments}

We evaluate FPPF on dynamical systems which highlight distinct
challenges in nonlinear filtering: scaling to high state dimension,
non-Gaussian posteriors that break Gaussian-assumption filters, and
long-horizon stability under temporally sparse observations.

\paragraph{Setup.}
ML-based DA methods are typically trained on simulated trajectories from the exact
dynamics \citep{rozet2023scorebased, chen2025flowdas}, but in
realistic settings this is rarely available; dynamics are known only up to model
error from unresolved physics, discretization, etc. We mirror this by
perturbing the training dynamics with process noise $\sigma_\eta$, simulating
$\bx_t = \mathbf{f}(\bx_{t-1}) + \sigma_\eta \boldsymbol{\zeta}_t$,
$\boldsymbol{\zeta}_t \sim \cN(\mathbf{0}, \bI)$, similar to \citet{tang2026score}; this also improves robustness at
inference. Evaluation uses the unperturbed dynamics, which fixes a single
ground-truth trajectory to score against. We obtain the
ground-truth initial state $\bx_0^\star$ by running the deterministic dynamics
through a burn-in from a random start, and center the initial ensemble at the truth,
$\bx_0^{(i)} \sim \cN(\bx_0^\star, \sigma_{\text{init}}^2 \bI)$, where
$\sigma_{\text{init}}$ is the per-coordinate standard deviation of the state
estimated from the training trajectories. For FlowDAS \citep{chen2025flowdas}, which
tracks a single trajectory rather than an ensemble, we initialize at $\bx_0^\star$.

\paragraph{Baselines.}
We compare against \textbf{BPF} \citep{gordon1993novel}
(transition-prior proposal), \textbf{APF} \citep{pitt1999auxiliary}
(one-step observation look-ahead), \textbf{EnKF}
\citep{evensen1994enkf} (dominant operational DA method), and
\textbf{FlowDAS} \citep{chen2025flowdas} (a recent stochastic-interpolant
DA method with observation-based guidance, outperforming prior generative
methods \citep{rozet2023scorebased}). We also include two
learned-proposal methods: \textbf{NASMC} \citep{gu2015nasmc}, with
diagonal-Gaussian and $K$-component MDN variants trained online via
particle-weighted log-likelihood; and \textbf{InfNN}
\citep{paige2016inference}, an autoregressive conditional density
(RNADE) trained offline by maximum likelihood. The other learned-proposal methods, NASMC and InfNN, are deployed with the same
SIR particle filter settings as FPPF for fairness. For architectural parity, the same backbone with matched parameter count is
used as the velocity field in FPPF, the drift network in FlowDAS, and the
conditional network in NASMC and InfNN; only the output head differs for
NASMC and InfNN, to match their parametric density families. Architecture choices and hyperparameters are
detailed in Appendix~\ref{app:architectures}; baseline implementation
details are in Appendix~\ref{app:baselines}.

\paragraph{Dynamical systems.}
We evaluate on three chaotic systems, each chosen to stress a different
filtering regime. Lorenz-63 \citep{lorenz1963deterministic} is a
three-dimensional model of atmospheric convection whose two-lobe attractor
produces multimodal forecast distributions under sparse observations,
isolating the non-Gaussian regime where Kalman-type updates fail.
Lorenz-96 \citep{lorenz1995predictability} is a spatially extended model on
a periodic latitude circle whose dimension can be set freely, letting us
probe scaling and localization in high dimensions. The
Kuramoto-Sivashinsky equation \citep{kuramoto1978, sivashinsky1977} is a fourth-order nonlinear
PDE exhibiting spatiotemporal chaos, which we use to test long-horizon
stability under temporally sparse observations. Full dynamics,
discretization, and observation operators for each system are given in the
corresponding subsections below.

\paragraph{Evaluation metrics.}
We report root mean square error (RMSE) for state estimation accuracy
and continuous ranked probability score (CRPS) \citep{matheson1976scoring} for probabilistic
calibration. CRPS measures the quality of probabilistic forecasts by
comparing the predicted CDF to the observed outcome; lower values
indicate better calibration. For an ensemble of size N, at time step $t$, it is defined as:
\begin{equation}
    \text{CRPS}_{t} = \frac{1}{N} \sum_{i=1}^N |x_{t}^{(i)} - x_{t}^\star| - \frac{1}{2N^2} \sum_{i=1}^N \sum_{i'=1}^N |x_{t}^{(i)} - x_{t}^{(i')}|.
    \label{eq:ess}
\end{equation}
We also monitor effective sample size
(ESS) over time to diagnose filter health,
$\mathrm{ESS}_t = 1 / \sum_{i=1}^{N} (w_t^{(i)})^2$, where $w_t^{(i)}$
denotes the normalized importance weight of particle $i$ at time $t$.
When the weights are uniform ($w_t^{(i)} = 1/N$), ESS attains its maximum value $N$;
an ESS near 1 indicates weight collapse.

\subsection{Scaling with state dimension}
\label{sec:experiments:scaling}

\subsubsection{Global filters across dimensions}

We test scaling on the Lorenz-96 system \citep{75462}, which models
atmospheric dynamics along a latitude circle. The state
$\bx_t \in \bR^{d_x}$ evolves according to
\begin{equation*}
    \frac{dx_j}{dt} = (x_{j+1} - x_{j-2})x_{j-1} - x_j + F + \xi_{j,t},
\end{equation*} 
with periodic boundary conditions and forcing $F = 8$ (chaotic regime).
Process noise is $\xi_{j,t} \sim \cN(0, \sigma_{\mathrm{proc}}^2)$
with $\sigma_{\mathrm{proc}} = 0.2$ and integration step
$\Delta t = 0.05$. We vary $d_x \in \{5, 10, 15, 20, 25, 50\}$ to create
systems with increasing state dimensionality.
Observations are $\bo_t = \arctan(\bx_t) + \bv_t$ with
$\bv_t \sim \cN(\mathbf{0}, \sigma_{\mathrm{obs}}^2 \bI)$,
$\sigma_{\mathrm{obs}} = 0.2$, applied at every integration step.

\begin{figure}[tb]
  \centering
  \includegraphics[width=\linewidth]{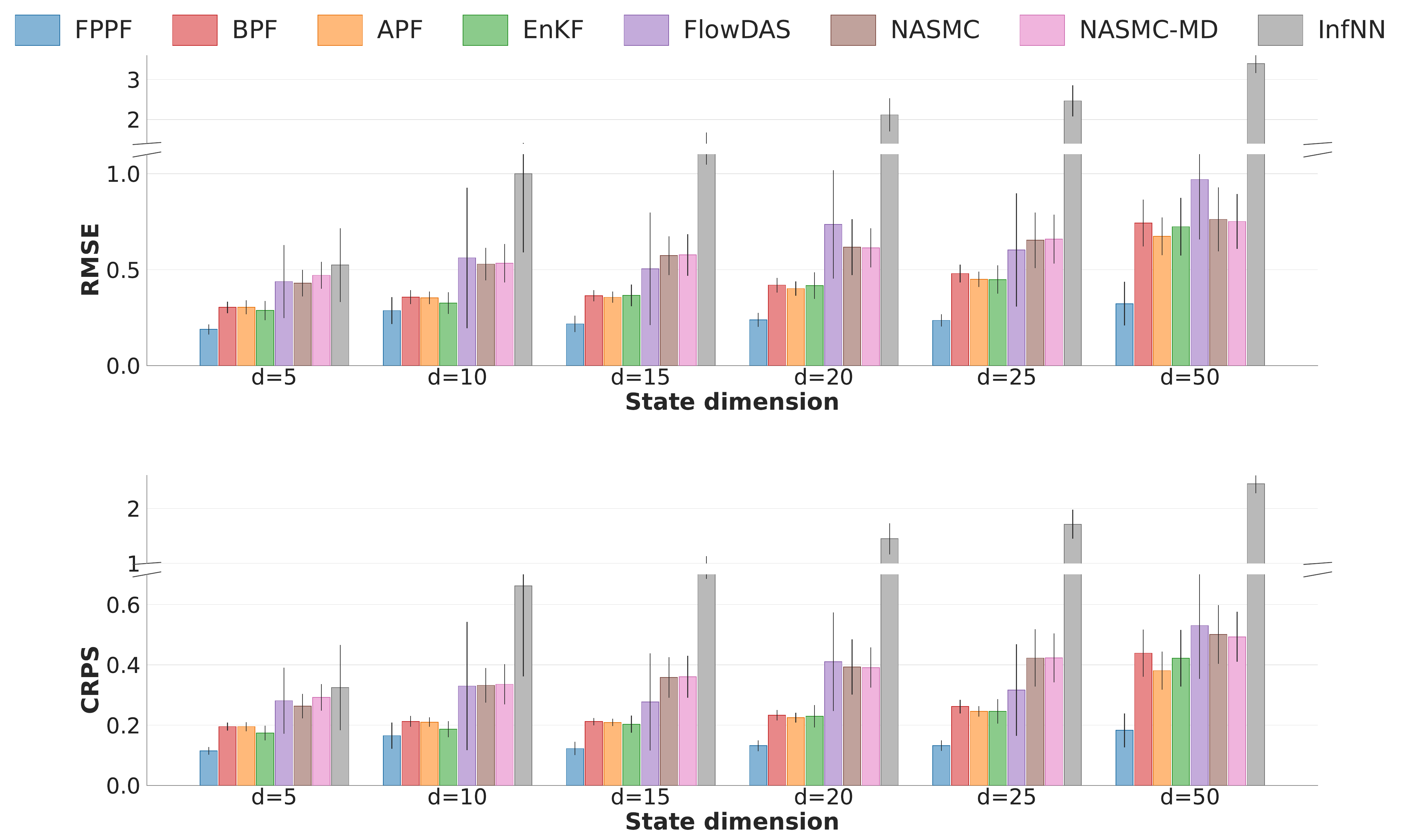}
  \caption{Scaling on Lorenz-96 with $\boldsymbol{\arctan(\bx)}$ observation operator across state
  dimensions $d_x \in \{5, 10, 15, 20, 25, 50\}$. FPPF attains the
  lowest RMSE and CRPS at every $d_x$, with the gap over baselines
  widening as dimension grows. Bars show means over 100 trajectories
  of 200 time steps and error bars denote one standard deviation. 
  Table in Appendix~\ref{app:lorenz96_full}.}
  \label{fig:lorenz96_scaling}
  \vspace{-1em}
\end{figure}

\textbf{Results.}  Figure~\ref{fig:lorenz96_scaling} reports results averaged over 100
trajectories of 200 steps, with $N = 1000$ particles for
$d_x \in \{5, 10\}$ and $N = 5000$ for $d_x \in \{15, 20, 25, 50\}$;
refer to Appendix~\ref{app:lorenz96_full} for the corresponding table. FPPF achieves the best RMSE and CRPS at every state dimension, and
the margin over BPF widens with $d_x$. Its learned proposal directly targets high-posterior regions by conditioning on observations, whereas the bootstrap proposal uses no observation information and increasingly suffers from the curse of dimensionality. APF, which incorporates observations through auxiliary resampling weights rather than the proposal itself, performs worse than FPPF, but shows gains over BPF that grow with dimension. NASMC also learns an observation-informed proposal but restricts it to a single Gaussian (NASMC-Gauss) or mixture density (NASMC-MD), and underperforms BPF, as does InfNN. EnKF performs slightly better than BPF: under
the near-Gaussian likelihood induced by arctan, the Gaussian update used in EnKF remains adequate.

\begin{figure}[t]
    \centering
    \includegraphics[width=\textwidth]{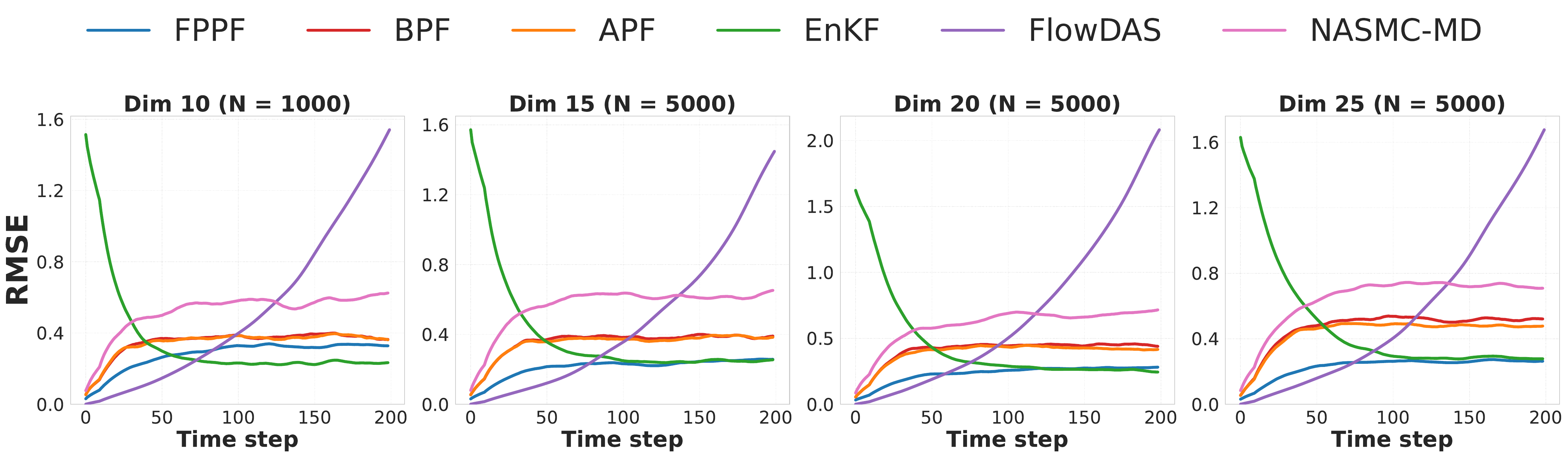}
    \caption{RMSE over time for Lorenz-96 at different state dimensions. FPPF (blue) maintains lower error than BPF. FlowDAS (purple) accumulates error rapidly due to its autoregressive generation and lack of principled update step. EnKF (green) has high error early due to poor initialization of sample covariance from the ensemble members.}
    \label{fig:lorenz96_time_rmse}
    \vspace{-1em}
\end{figure}

Figure~\ref{fig:lorenz96_time_rmse} plots RMSE over time for a subset of the baselines. Filtering
methods exhibit stable tracking throughout, with FPPF maintaining the
lowest error. FlowDAS \citep{chen2025flowdas} accumulates error rapidly due to its ``single particle'' autoregressive generation and lack of principled update step. EnKF has high error early due to poor initialization of sample covariance from the ensemble members.

\begin{figure}[htbp]
    \centering
    \includegraphics[width=\textwidth]{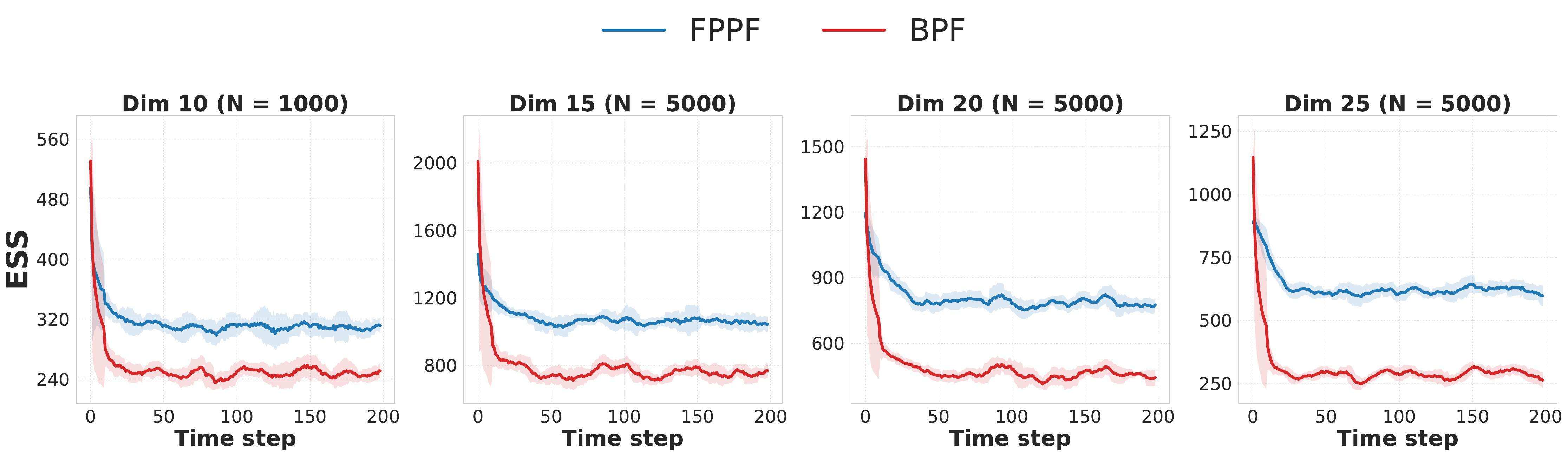}
    \caption{Pre-resample ESS over time for Lorenz-96 across state dimensions. Curves show rolling means (window 10) over 100 trajectories; shading indicates one standard deviation across trajectories. FPPF maintains a higher ESS than BPF, and the gap widens with $d_x$.}
    \label{fig:lorenz96_time_ess}
    \vspace{-1em}
\end{figure}

Figure~\ref{fig:lorenz96_time_ess} plots pre-resample ESS over time.
FPPF maintains a higher ESS than BPF, and the gap widens with
dimension. At $d_x = 25$ the absolute ESS is
nonetheless low enough to be concerning, and naively increasing the
particle count is computationally infeasible. This motivates the
localized variant L-FPPF (Section~\ref{sec:method2}, Section~\ref{sec:experiments:local}), which exploits
the local correlation structure of spatially extended systems to
mitigate degeneracy and scale efficiently with state dimensionality.

\subsubsection{Localized filters for high-dimensional systems.}
\label{sec:experiments:local}

We evaluate L-FPPF on Lorenz-96 at $d_x = 50$, $100$, $500$, and
$1000$, a regime where global particle filters collapse. The L-FPPF proposal
is trained once on the $d_x{=}25$ dataset and applied zero-shot at all
four target dimensions. Network inputs have the window dimension $2r{+}1$, where we fix localization radius $r{=}4$ across all
filtering methods. This allows us to train a lightweight model just once,
and the per-step sampling and likelihood integration stays tractable
even at $d_x{=}1000$. Localization also allows us to reduce ensemble size for all PFs to
$N_p{=}500$ particles, with negligible performance decline. While the
necessary $N_p$ for global PFs scale exponentially with
the effective state dimension
\citep{snyder2008obstacles, farchi2018comparison}, block-wise
localization replaces the global problem with overlapping local ones
whose effective dimension is set by the localization radius rather
than $d_x$, so the ensemble requirement stays bounded as the state
dimension grows \citep{poterjoy2016localized, farchi2018comparison}.

For evaluation, we introduce two localized
DA baselines. \textbf{LETKF}
\citep{hunt2007efficient} performs EnKF analysis in local windows
under a Gaussian-posterior assumption and is deployed in
operational settings
\citep{houtekamer2014higher,schraff2016kilometre}. \textbf{LBPF}
\citep{poterjoy2016localized,farchi2018comparison} propagates
particles under global dynamics with tapering local importance
weights and per-site resampling; it shares L-FPPF's block-wise
machinery and resampling step, isolating the contribution of the learned local proposal. We
also report global BPF, APF, and EnKF for reference. 

{\bf Results.} Figure~\ref{fig:lorenz96_localized_arctan} reports RMSE; the full table including CRPS is in Appendix~\ref{app:localized_tables}. L-FPPF achieves the lowest RMSE and CRPS at all dimensions, staying relatively flat as $d_x$ grows from 50 to 1000 while BPF, APF, and EnKF degrade sharply. LBPF is likewise stable with dimension. This is expected, since its block-wise weighting and resampling reduce the effective dimension of the importance-sampling problem from \( d_x \) to a local neighborhood of size set by \( r \). L-FPPF reduces RMSE over LBPF at every dimension. Both filters avoid global particle collapse, but the learned observation-informed proposal places particles nearer the current local posterior, further reducing the residual local weight variance.

\begin{figure}[t]
  \centering
  \includegraphics[width=\linewidth]{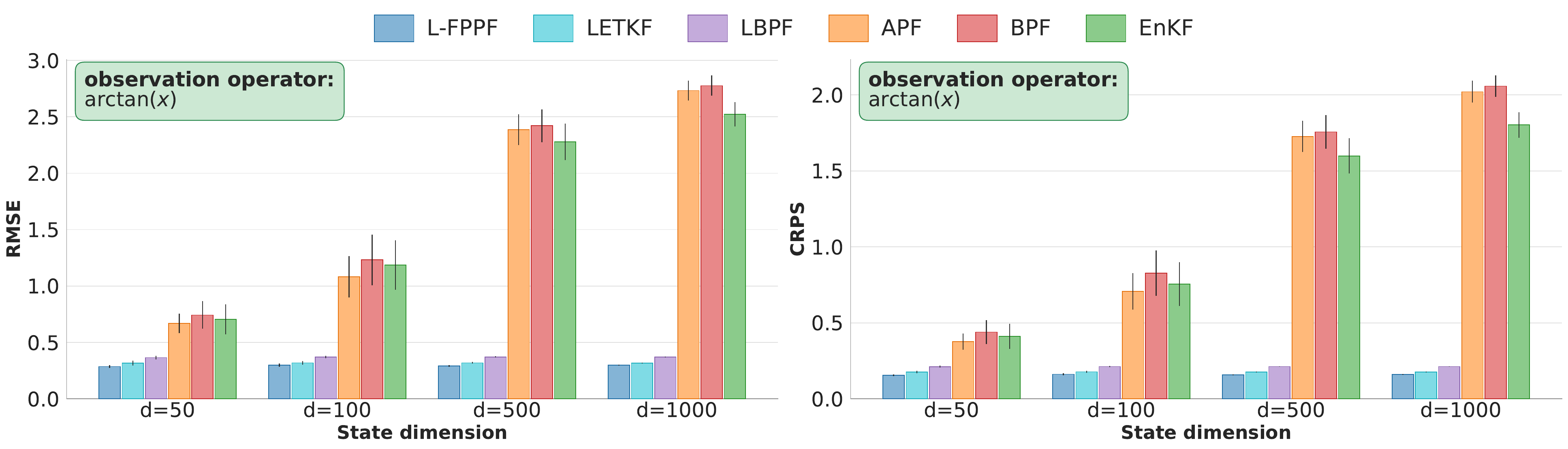}
  \caption{RMSE and CRPS on Lorenz-96 at $d_x \in \{50, 100, 500, 1000\}$ for localized and global filters under observation operator $h(\bx) = \arctan(\bx)$. L-FPPF is trained once on $d_x{=}25$ and applied zero-shot at all target dimensions; localization radius $r{=}4$ is fixed across L-FPPF, LBPF, and LETKF. Bars show mean RMSE across independent runs and error bars denote one standard deviation. Table is reported in Appendix~\ref{app:localized_tables}.}
  \label{fig:lorenz96_localized_arctan}
\end{figure}

\begin{figure}[t]
  \centering
  \includegraphics[width=\linewidth]{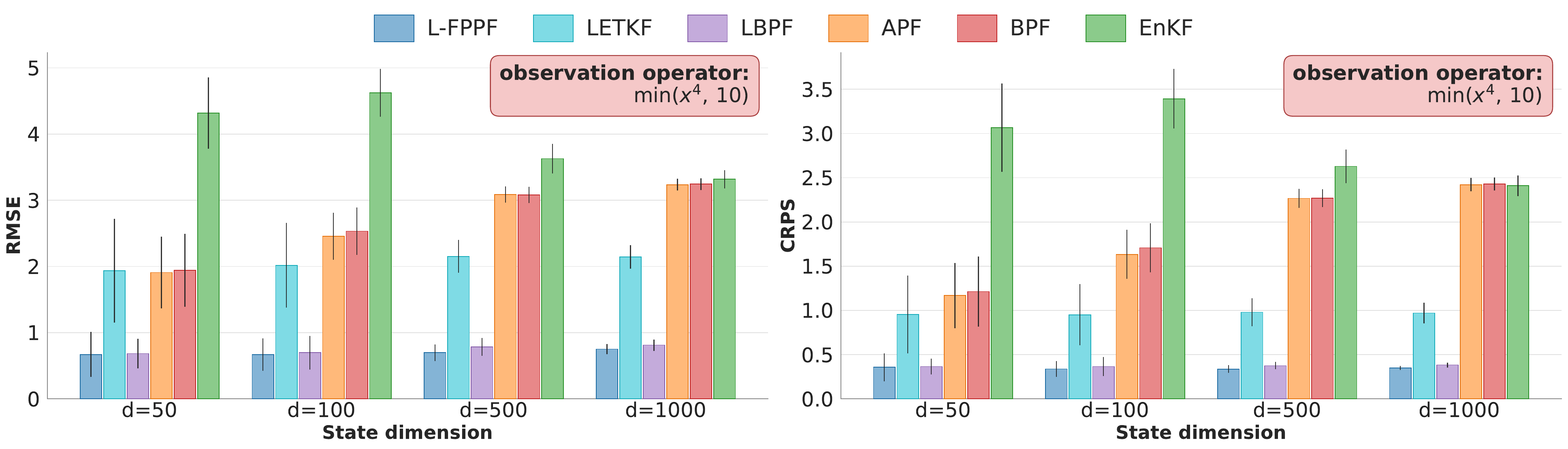}
  \caption{RMSE and CRPS on Lorenz-96 at $d_x \in \{50, 100, 500, 1000\}$ for localized and global filters under observation operator $h(\bx) = \min(\bx^4, 10)$. L-FPPF is trained once on $d_x{=}25$ and applied zero-shot at all target dimensions; localization radius $r{=}4$ is fixed across L-FPPF, LBPF, and LETKF. Bars show mean RMSE across independent runs and error bars denote one standard deviation. Table is reported in Appendix~\ref{app:localized_tables}.}
  \label{fig:lorenz96_localized_quadcapped}
\end{figure}

\subsection{Non-linear and non-Gaussian filtering regimes}
\label{sec:experiments:nongaussian}

EnKF and related Gaussian-assumption filters are common in operational data
assimilation, but their Gaussian posterior approximation breaks down
in two key regimes: (i) when the observation operator induces a
non-Gaussian observation distribution, and (ii) when the dynamics produce a
multimodal prior between observations. We construct experiments that
isolate each.

\paragraph{Non-Gaussian likelihoods.}
We retain the Lorenz-96 setup of Section~\ref{sec:experiments:scaling}
but replace the arctangent operator with $h(\bx) = \min(\bx^4, 10)$
elementwise. This is substantially more challenging: $\bx^4$ is sign-ambiguous,
inducing per-component multimodality, and the cap saturates for
$|x_j| > 10^{1/4}$, where the observation carries no magnitude information.
Table~\ref{tab:lorenz96_quad_capped_s02} reports the same scaling sweep.
EnKF fails catastrophically at all dimensions, since a Gaussian posterior
cannot represent the induced bimodality. APF also degrades, as its one-step
look-ahead is unreliable under a sign-ambiguous, saturating observation
model. NASMC-MD recovers some advantage over its Gaussian counterpart by
accommodating multimodality. FPPF retains the best RMSE and CRPS at every
dimension, with larger margins over BPF than under arctan---likely
reflecting the flow proposal's capacity to learn the multimodal
observation-conditional density directly from data.
Figure~\ref{fig:lorenz96_localized_quadcapped} extends the localized comparison
to $d_x \in \{50, 100, 500, 1000\}$ (full table in
Appendix~\ref{app:localized_tables}). LETKF inherits EnKF's failure mode
locally and degrades sharply; LBPF is competitive but still trails L-FPPF.

\paragraph{Multimodal distributions.}
Lorenz-63 \citep{lorenz63} is a three-dimensional chaotic model of atmospheric
convection whose attractor has two lobes between which trajectories switch at
irregular intervals. The bimodal regime this experiment targets emerges
under temporally sparse observations. Between updates the ensemble evolves
freely under the nonlinear dynamics, and members straddling a lobe transition
are carried into different lobes, so the forecast/prior at the next assimilation
time is non-Gaussian and bimodal \citep{anderson1999montecarlo}. Were
observations sufficiently frequent, each update would pin the ensemble to the truth within a
single lobe and the prior would stay unimodal, leaving little for a non-Gaussian
filter to exploit. We therefore assimilate only every 50 integration steps,
allowing substantial ensemble dispersion between updates. The state
$\bx_t = (x, y, z) \in \bR^3$ evolves as
\begin{equation*}
\dot{x} = \sigma(y - x), \qquad
\dot{y} = x(\rho - z) - y, \qquad
\dot{z} = xy - \beta z,
\end{equation*}
with $\sigma = 10$, $\rho = 28$, $\beta = 8/3$, integrated at $\Delta t = 0.01$
with process noise $\sigma_{\mathrm{proc}} = 0.25$. Observations are
$\bo_t = \arctan(\bx_t) + \bv_t$ with
$\bv_t \sim \cN(\mathbf{0}, \sigma_{\mathrm{obs}}^2 \bI)$ and
$\sigma_{\mathrm{obs}} = 0.25$.

Table~\ref{tab:l63_freq50_results} reports results over 50 trajectories of 1000
steps with $N = 1000$ particles. FlowDAS is the worst by a wide margin,
consistent with the autoregressive error accumulation for long trajectories in previous experiments. 
Among the remaining filters EnKF is the weakest, its
Gaussian posterior is unable to capture the bimodal filtering density, which is
precisely the failure this setup isolates. BPF, APF, and FPPF are essentially
tied. The bootstrap proposal is adequate here only because the three-dimensional
state keeps particle weight variance small, so weights do not
collapse (Section~\ref{sec:prelims:degeneracy}); this is the low-dimensional regime in
which particle filters have always been reliable, and the benefit of an
observation-informed proposal instead grows with state dimension
(Section~\ref{sec:experiments:scaling}).

\begin{table*}[t]
\vskip 0.1in
\centering
\small
\setlength{\tabcolsep}{6pt}
\begin{tabular}{@{}lccccc@{}}
\toprule
Metric & BPF & APF & EnKF & FPPF & FlowDAS \\
\midrule
RMSE $\downarrow$ 
& \textbf{2.7473} {\scriptsize $\pm$ 0.4553}
& 2.7650 {\scriptsize $\pm$ 0.4490}
& 4.7246 {\scriptsize $\pm$ 0.6827}
& 2.7504 {\scriptsize $\pm$ 0.4436}
& {7.2579 \scriptsize $\pm$ 0.9512} \\
CRPS $\downarrow$ 
& 1.6760 {\scriptsize $\pm$ 0.2662}
& 1.6839 {\scriptsize $\pm$ 0.2617}
& 3.0084 {\scriptsize $\pm$ 0.4655}
& \textbf{1.6755} {\scriptsize $\pm$ 0.2527}
& 4.9863 {\scriptsize $\pm$ 0.7024} \\
\bottomrule
\end{tabular}
\caption{Lorenz-63 results with sparse observations every 50 time steps, observation noise $\sigma_\text{obs}$ = 0.25, for 1000 length trajectories.}
\label{tab:l63_freq50_results}
\vskip -0.1in
\end{table*}

\subsection{Long-horizon stability under temporally sparse observations}
\label{sec:experiments:ks}

Operational DA systems must remain stable over long trajectories with
observations that are sparse in time. Between observations, the system
evolves substantially, the prior at the next observation step can
become highly non-Gaussian, and small errors compound across the
rollout. We use the Kuramoto-Sivashinsky (KS) equation, a fourth-order
nonlinear PDE exhibiting spatiotemporal chaos, to test this regime:
\begin{equation}
\partial_t u = -u\, \partial_x u - \partial_x^2 u - \partial_x^4 u,
\end{equation}
with periodic boundary conditions on $[0, L]$, $L \in \{16\pi, 32\pi\}$. We
discretize using a pseudo-spectral method with $J = 128$ grid points (so
$d_x = J$), time step $\Delta t = 0.25$, and process noise
$\sigma_{\mathrm{proc}} = 0.1$. Linearizing about the trivial state $u \equiv 0$,
each Fourier mode grows at rate $\lambda_k = k^2 - k^4$, so wavenumbers
$0 < |k| < 1$ are linearly unstable. The admissible wavenumbers are
$k = 2\pi n / L$, so a larger domain places more of them in the unstable band;
$L = 16\pi$ admits $7$ unstable modes and $L = 32\pi$ admits $15$. $L = 32\pi$ is thus the
more chaotic, complex regime. All grid points are observed
through
\begin{equation}
\bo_t = h(\bx_t) + \bv_t, \qquad
\bv_t \sim \cN(\mathbf{0}, \sigma_{\mathrm{obs}}^2 \bI), \quad
\sigma_{\mathrm{obs}} = 0.1,
\end{equation}
where $h$ is applied elementwise and is either the smooth $\arctan$ operator or
the sign-ambiguous, saturating $\min(x^4, 10)$ operator, and observations
arrive every 10 steps.

Table~\ref{tab:ks_results} reports results over 25 trajectories of 1000 steps
using $N = 5000$ particles, for both observation operators. FlowDAS fails to
track the system over the full trajectory in all settings. 
Under the smooth $\arctan$ operator, EnKF achieves the lowest
error on both domains, with FPPF close behind, since the Gaussian observation update
is not heavily violated. Under the more difucult $\min(x^4, 10)$
operator, FPPF is best on both $L = 16\pi$ and
$L = 32\pi$. On the more chaotic $L = 32\pi$ domain FPPF
improves the most substantially over BPF. Another proposal learning method, NASMC, also performs well. EnKF
performs the worst
since its Gaussian likelihood assumption now breaks down under the multimodal
observation distribution.

\newcommand{\stderr}[1]{{\fontsize{6}{7}\selectfont $\pm$ #1}}
\begin{table*}[t]
\vspace{-1em}
\vskip 0.1in
\centering
\footnotesize
\setlength{\tabcolsep}{3pt}
\begin{tabular}{@{}lcccccccc@{}}
\toprule
& \multicolumn{4}{c}{$L = 16\pi$}
& \multicolumn{4}{c}{$L = 32\pi$} \\
\cmidrule(lr){2-5}\cmidrule(l){6-9}
& \multicolumn{2}{c}{$\arctan$}
& \multicolumn{2}{c}{$\min(z^4,10)$}
& \multicolumn{2}{c}{$\arctan$}
& \multicolumn{2}{c}{$\min(z^4,10)$} \\
\cmidrule(lr){2-3}\cmidrule(lr){4-5}\cmidrule(lr){6-7}\cmidrule(l){8-9}
Method
& RMSE $\downarrow$ & CRPS $\downarrow$
& RMSE $\downarrow$ & CRPS $\downarrow$
& RMSE $\downarrow$ & CRPS $\downarrow$
& RMSE $\downarrow$ & CRPS $\downarrow$ \\
\midrule
BPF
& 0.14 \stderr{0.017} & 0.08 \stderr{0.011}
& 0.91 \stderr{0.471} & 0.65 \stderr{0.373}
& 1.21 \stderr{0.091} & 0.84 \stderr{0.073}
& 1.59 \stderr{0.055} & 1.17 \stderr{0.057} \\
APF
& 0.11 \stderr{0.004} & 0.06 \stderr{0.002}
& 0.49 \stderr{0.478} & 0.33 \stderr{0.364}
& 0.76 \stderr{0.197} & 0.48 \stderr{0.141}
& 1.53 \stderr{0.101} & 1.12 \stderr{0.094} \\
EnKF
& \textbf{0.07} \stderr{0.002} & \textbf{0.05} \stderr{0.001}
& 0.18 \stderr{0.380} & 0.12 \stderr{0.279}
& \textbf{0.11} \stderr{0.002} & \textbf{0.06} \stderr{0.001}
& 1.65 \stderr{0.107} & 1.17 \stderr{0.116} \\
NASMC-Gauss
 & \underline{0.09} \stderr{0.004} & 0.05 \stderr{0.001}
 & \underline{0.15} \stderr{0.056} & \underline{0.08} \stderr{0.033}
 & \underline{0.19} \stderr{0.006} & \underline{0.10} \stderr{0.003}
 & \underline{1.24} \stderr{0.220} & \underline{0.82} \stderr{0.175} \\
NASMC-MD
 & \underline{0.09} \stderr{0.003} & 0.05 \stderr{0.001}
 & \underline{0.15} \stderr{0.081} & 0.09 \stderr{0.050}
 & \underline{0.19} \stderr{0.007} & \underline{0.10} \stderr{0.004}
 & \underline{1.24} \stderr{0.220} & \underline{0.82} \stderr{0.175} \\
InfNN
& 1.48 \stderr{0.102} & 1.08 \stderr{0.082}
& 1.84 \stderr{0.084} & 1.44 \stderr{0.076}
& 1.52 \stderr{0.026} & 1.12 \stderr{0.022}
& 1.77 \stderr{0.049} & 1.36 \stderr{0.042} \\
FPPF
& \underline{0.09} \stderr{0.003} & \underline{0.05} \stderr{0.001}
& \textbf{0.11} \stderr{0.013} & \textbf{0.06} \stderr{0.006}
& \underline{0.19} \stderr{0.009} & \underline{0.10} \stderr{0.004}
& \textbf{1.15} \stderr{0.231} & \textbf{0.76} \stderr{0.186} \\
FlowDAS
& 1.18 \stderr{0.067} & 0.70 \stderr{0.038}
& 1.55 \stderr{0.110} & 1.12 \stderr{0.090}
& 1.23 \stderr{0.016} & 0.72 \stderr{0.008}
& 1.62 \stderr{0.060} & 1.18 \stderr{0.055} \\
\bottomrule
\end{tabular}
\caption{Kuramoto-Sivashinsky results ($J=128$) with sparse observations every 10 steps, observation noise $\sigma_\text{obs}=0.1$, and two observation operators on domains $L=16\pi$ and $L=32\pi$. Best results in \textbf{bold}, second best \underline{underlined}.}
\label{tab:ks_results}
\end{table*}

\subsection{Additional experiments}
\label{sec:experiments:additional}

\paragraph{Convergence for linear-Gaussian setting.}
\label{sec:experiments:lg_sanity}
To verify that the rectified flow proposal converges to the correct target, we test on a synthetic linear Gaussian state-space model (full settings in Appendix~\ref{app:lg_sanity}), where the locally optimal one-step proposal $q^\star(\bx_t \mid \bx_{t-1}, \bo_t)$ admits a closed form. Using the same architecture as our L96 experiments, we train $q_\phi$ on a $d_x{=}8$ system for 200 epochs. 

\begin{figure}[htbp]
\vspace{-1em}
    \centering
    \includegraphics[width=0.66\textwidth]{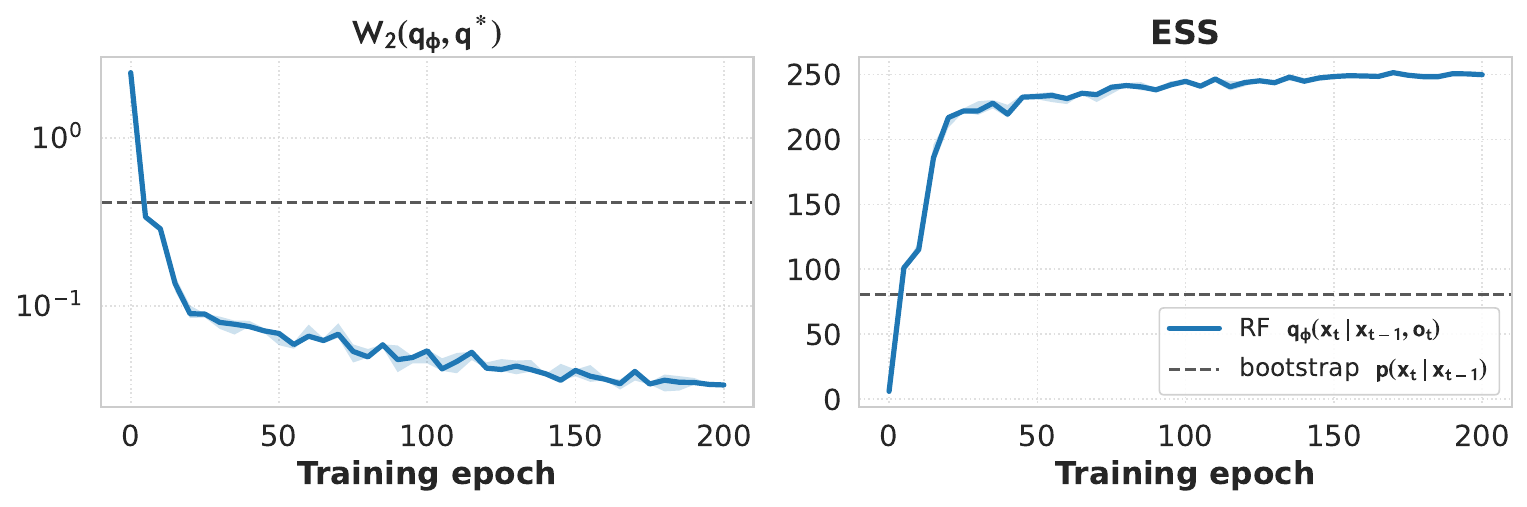}
    \hfill
    \includegraphics[width=0.33\textwidth]{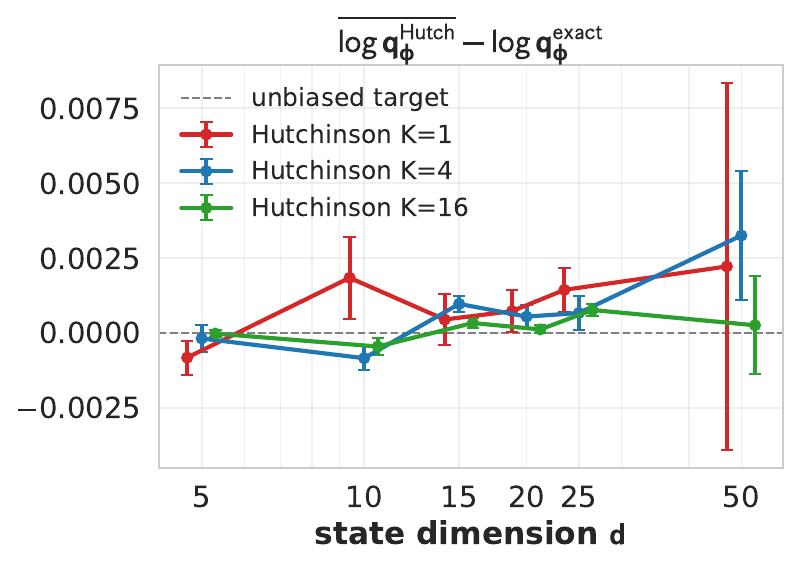}
    \caption{\textbf{Left:} Wasserstein $W_2$ distance from the learned proposal $q_\phi$ to
    the closed-form optimal proposal $q^\star$ over training epochs, averaged over
    $500$ test conditioning pairs; it decreases toward $0$, while the dashed line is
    the bootstrap proposal's $W_2$ to $q^\star$.
    \textbf{Middle:} one-step ESS out of $N{=}250$ over training epochs; $q_\phi$
    rises toward the maximum of $250$, while the bootstrap proposal's (dashed) is near $80$.
    \textbf{Right:} bias of the Hutchinson trace estimator relative to the exact
    trace on the Lorenz-96 proposals across state dimension $d_x$, for
    $K \in \{1, 4, 16\}$ probes; points show mean bias where ideal is 0. Error bars indicate standard
    deviation across $N_{\mathrm{pairs}}{=}10$ conditioning pairs.}
    \label{fig:ablation}
    \vspace{-1em}
\end{figure}

The left panel of Figure~\ref{fig:ablation} reports the Wasserstein $W_2$ distance between $q_\phi$ and $q^\star$, averaged over $500$ test set conditioning pairs $(\bx_{t-1}, \bo_t)$. It decreases over training to 0, indicating convergence to the closed-form optimal proposal. The dotted horizontal line indicates the $W_2$ distance between the bootstrap proposal $p(\bx_t \mid \bx_{t-1})$ and the closed-form optimal proposal $q^\star(\bx_t \mid \bx_{t-1}, \bo_t)$. 

The middle panel reports a one-step ESS. For each conditioning pair we draw
$N{=}250$ samples $\bx_t^{(i)} \sim q_\phi(\cdot \mid \bx_{t-1}, \bo_t)$, form the
one-step SIR importance weights
\begin{equation*}
    \tilde w^{(i)} \propto
    \frac{\prob(\bo_t \mid \bx_t^{(i)})\,\prob(\bx_t^{(i)} \mid \bx_{t-1})}
         {q_\phi(\bx_t^{(i)} \mid \bx_{t-1}, \bo_t)},
\end{equation*}
and compute $\mathrm{ESS} = 1 / \sum_i (\bar w^{(i)})^2$ from the normalized weights
$\bar w^{(i)}$, averaging over the $500$ pairs. This is the ESS the filter would
achieve after a single observation update with the proposal. It attains its
ceiling of $N{=}250$ when the incremental weights are uniform
($\bar w^{(i)} = 1/N$), which is exactly what the optimal proposal achieves,
since under $q^\star$ the incremental weight is independent of the sampled
$\bx_t$ (Eq.~\eqref{eq:opt-weight}). The bootstrap baseline collapses to near
$80$ (dashed line), while $q_\phi$ stabilizes near $250$ over training.

\paragraph{Hutchinson trace estimator.}
Computing $\log q_\phi(\bx_t \mid \bx_{t-1}, \bo_t)$ requires the divergence of
the proposal velocity field, which
we estimate with the Hutchinson trace estimator
(Section~\ref{sec:method:weights}, Eq.~\eqref{eq:weight_update}). One concern is that the
single-probe estimator ($K{=}1$) injects noise into $\log q_\phi$ and could
inflate importance weight variance. We quantify the single-probe estimator error on the trained Lorenz-96 RF proposals from Section~\ref{sec:experiments:scaling}. At $N_{\mathrm{pairs}}{=}10$ test-set conditioning pairs we draw $\bx_t \sim q_\phi$,
evaluate the Hutchinson estimate of $\log q_\phi$ at $K \in \{1, 4, 16\}$ probes
($M{=}200$ random-probe seeds each), and measure its bias and standard deviation
against the deterministic exact-trace value. The empirical bias is
consistent with zero at every $d_x$ within pair-to-pair, see
Figure~\ref{fig:ablation} (right). Estimator standard deviation stays below $0.05$ for $d_x \leq 25$ but grows to $0.247$ at $d_x{=}50$ for $K{=}1$; it is substantially reduced for $K=16$. While we keep $K{=}1$ by default for efficiency, at high dimensions increasing $K$ at the expense of compute may improve performance. 

\section{Conclusion and Future Work}
\label{sec:conclusion}

We introduced Flow Proposal Particle Filters (FPPF), which replace the bootstrap dynamics with a learned conditional flow proposal approximating the variance-minimizing optimal proposal $p(\bx_t \mid \bx_{t-1}, \bo_t)$. Conditional flow matching provides efficient sampling and tractable density evaluation, so the SIR weight update remains correct rather than absorbed into approximate guidance. The localized variant L-FPPF uses a patch-based velocity network, making the SIR correction itself decomposable. Across chaotic systems with non-Gaussian likelihoods and long horizons, FPPF and L-FPPF improve state estimation and calibration over classical, learned-proposal, and autoregressive generative baselines, and L-FPPF generalizes to high dimensions zero-shot at fixed ensemble size. 

Several directions remain open. A consequential challenge is scaling to operational systems such as ERA5-scale reanalysis \citep{Dee2011ERAInterim}. Localization is encouraging in this regard, as it caps the effective dimension at the localization radius rather than the ambient state dimension, allowing methods like L-FPPF to operate stably at large scale with fixed ensemble size. Further, in practice, atmospheric observations are spatially sparse, irregularly distributed, and collected from heterogeneous instruments, with complex, nonlinear observation operators mapping between state and measurement spaces. These features introduce additional structure absent from our benchmarks and can significantly complicate inference.
Second, since FPPF assumes access to the transition density for the
SIR weight, extending it to data-driven dynamics through a learned forecast
emulator is a natural next step. A third concerns cost, as sampling and
likelihood evaluation each require multiple integration steps per particle per
step; few-step distillation \citep{song2023consistency, geng2025meanflow,
frans2024shortcut} together with fast divergence estimation \citep{ai2026joint}
could reduce this (Appendix~\ref{app:baselines}). Finally, FPPF targets the
proposal rather than the analysis distribution, and combining it with latent or
score-based ensemble representations is a promising direction we leave to future
work.

\bibliographystyle{abbrvnat}
\bibliography{references}

\begin{thebibliography}{93}
\providecommand{\natexlab}[1]{#1}
\providecommand{\url}[1]{\texttt{#1}}
\expandafter\ifx\csname urlstyle\endcsname\relax
  \providecommand{\doi}[1]{doi: #1}\else
  \providecommand{\doi}{doi: \begingroup \urlstyle{rm}\Url}\fi

\bibitem[Ades and van Leeuwen(2015)]{ades2015ewpf}
M.~Ades and P.~J. van Leeuwen.
\newblock The equivalent-weights particle filter in a high-dimensional system.
\newblock \emph{Quarterly Journal of the Royal Meteorological Society}, 141\penalty0 (687):\penalty0 484--503, 2015.
\newblock \doi{10.1002/qj.2370}.

\bibitem[Ai et~al.(2026)Ai, He, Gu, Salakhutdinov, Kolter, Boffi, and Simchowitz]{ai2026joint}
X.~Ai, Y.~He, A.~Gu, R.~Salakhutdinov, J.~Z. Kolter, N.~M. Boffi, and M.~Simchowitz.
\newblock Joint distillation for fast likelihood evaluation and sampling in flow-based models.
\newblock In \emph{The Fourteenth International Conference on Learning Representations (ICLR)}, 2026.
\newblock URL \url{https://openreview.net/forum?id=8uZ5UdIul2}.
\newblock arXiv:2512.02636.

\bibitem[Anderson and Moore(1979)]{anderson1979optimal}
B.~D.~O. Anderson and J.~B. Moore.
\newblock \emph{Optimal Filtering}.
\newblock Prentice-Hall, Englewood Cliffs, NJ, 1979.

\bibitem[Anderson and Anderson(1999)]{anderson1999montecarlo}
J.~L. Anderson and S.~L. Anderson.
\newblock A {M}onte {C}arlo implementation of the nonlinear filtering problem to produce ensemble assimilations and forecasts.
\newblock \emph{Monthly Weather Review}, 127\penalty0 (12):\penalty0 2741--2758, 1999.
\newblock \doi{10.1175/1520-0493(1999)127<2741:AMCIOT>2.0.CO;2}.

\bibitem[Andrae et~al.(2026)Andrae, Larsson, Takao, Landelius, and Lindsten]{andrae2026daisi}
M.~Andrae, E.~Larsson, S.~Takao, T.~Landelius, and F.~Lindsten.
\newblock Daisi: Data assimilation with inverse sampling using stochastic interpolants.
\newblock In \emph{Proceedings of the 43rd International Conference on Machine Learning (ICML)}, 2026.
\newblock URL \url{https://arxiv.org/abs/2512.00252}.

\bibitem[Arulampalam et~al.(2002)Arulampalam, Maskell, Gordon, and Clapp]{arulampalam2002tutorial}
M.~S. Arulampalam, S.~Maskell, N.~Gordon, and T.~Clapp.
\newblock A tutorial on particle filters for online nonlinear/non-{Gaussian} {Bayesian} tracking.
\newblock \emph{IEEE Transactions on Signal Processing}, 50\penalty0 (2):\penalty0 174--188, 2002.
\newblock \doi{10.1109/78.978374}.

\bibitem[Bao et~al.(2024)Bao, Zhang, and Zhang]{bao_ensemble_2024}
F.~Bao, Z.~Zhang, and G.~Zhang.
\newblock An ensemble score filter for tracking high-dimensional nonlinear dynamical systems.
\newblock \emph{Computer Methods in Applied Mechanics and Engineering}, 432:\penalty0 117447, Dec. 2024.
\newblock ISSN 0045-7825.
\newblock \doi{10.1016/j.cma.2024.117447}.

\bibitem[Bar-Shalom et~al.(2001)Bar-Shalom, Li, and Kirubarajan]{BarShalom2001TrackingNavigation}
Y.~Bar-Shalom, X.~R. Li, and T.~Kirubarajan.
\newblock \emph{Estimation with Applications to Tracking and Navigation: Theory, Algorithms and Software}.
\newblock John Wiley \& Sons, New York, 2001.
\newblock ISBN 9780471416555.
\newblock \doi{10.1002/0471221279}.

\bibitem[Bengtsson et~al.(2008)Bengtsson, Bickel, and Li]{Bengtsson_2008}
T.~Bengtsson, P.~Bickel, and B.~Li.
\newblock \emph{Curse-of-dimensionality revisited: Collapse of the particle filter in very large scale systems}, page 316–334.
\newblock Institute of Mathematical Statistics, 2008.
\newblock ISBN 0940600749.
\newblock \doi{10.1214/193940307000000518}.
\newblock URL \url{http://dx.doi.org/10.1214/193940307000000518}.

\bibitem[Bishop et~al.(2001)Bishop, Etherton, and Majumdar]{bishop2001adaptive}
C.~H. Bishop, B.~J. Etherton, and S.~J. Majumdar.
\newblock Adaptive sampling with the ensemble transform {K}alman filter. {P}art {I}: {T}heoretical aspects.
\newblock \emph{Monthly Weather Review}, 129\penalty0 (3):\penalty0 420--436, 2001.
\newblock \doi{10.1175/1520-0493(2001)129<0420:ASWTET>2.0.CO;2}.

\bibitem[Bocquet et~al.(2020)Bocquet, Brajard, Carrassi, and Bertino]{bocquet2020bayesian}
M.~Bocquet, J.~Brajard, A.~Carrassi, and L.~Bertino.
\newblock {B}ayesian inference of chaotic dynamics by merging data assimilation, machine learning and expectation-maximization.
\newblock \emph{Foundations of Data Science}, 2\penalty0 (1):\penalty0 55--80, 2020.
\newblock \doi{10.3934/fods.2020004}.

\bibitem[Brajard et~al.(2020)Brajard, Carrassi, Bocquet, and Bertino]{brajard2020combining}
J.~Brajard, A.~Carrassi, M.~Bocquet, and L.~Bertino.
\newblock Combining data assimilation and machine learning to emulate a dynamical model from sparse and noisy observations: A case study with the {L}orenz 96 model.
\newblock \emph{Journal of Computational Science}, 44:\penalty0 101171, 2020.
\newblock \doi{10.1016/j.jocs.2020.101171}.

\bibitem[Branchini and Elvira(2020)]{Branchini2020OptimizedAP}
N.~Branchini and V.~Elvira.
\newblock Optimized auxiliary particle filters: adapting mixture proposals via convex optimization.
\newblock In \emph{Conference on Uncertainty in Artificial Intelligence}, 2020.
\newblock URL \url{https://api.semanticscholar.org/CorpusID:235446306}.

\bibitem[Chen et~al.(2025)Chen, Jia, Qu, Sun, and Fessler]{chen2025flowdas}
S.~Chen, Y.~Jia, Q.~Qu, H.~Sun, and J.~A. Fessler.
\newblock Flow{DAS}: A stochastic interpolant-based framework for data assimilation.
\newblock In \emph{The Thirty-ninth Annual Conference on Neural Information Processing Systems}, 2025.
\newblock URL \url{https://openreview.net/forum?id=1nWqhiulqD}.

\bibitem[Chen and Li(2023)]{chen2023overview}
X.~Chen and Y.~Li.
\newblock An overview of differentiable particle filters for data-adaptive sequential {B}ayesian inference.
\newblock \emph{Foundations of Data Science}, 2023.
\newblock \doi{10.3934/fods.2023014}.

\bibitem[Chen and Li(2024)]{chen2024nfdpf}
X.~Chen and Y.~Li.
\newblock Normalizing flow-based differentiable particle filters.
\newblock \emph{IEEE Transactions on Signal Processing}, 2024.
\newblock arXiv:2403.01499.

\bibitem[Chen et~al.(2022)Chen, Sanz-Alonso, and Willett]{chen2023autodifferentiable}
Y.~Chen, D.~Sanz-Alonso, and R.~Willett.
\newblock Autodifferentiable ensemble {K}alman filters.
\newblock \emph{SIAM Journal on Mathematics of Data Science}, 4\penalty0 (2):\penalty0 801--833, 2022.
\newblock \doi{10.1137/21M1434477}.

\bibitem[Chen(2003)]{chen2003bayesian}
Z.~Chen.
\newblock Bayesian filtering: From kalman filters to particle filters, and beyond.
\newblock \emph{Statistics}, 182\penalty0 (1):\penalty0 1--69, 2003.
\newblock \doi{10.1080/02331880309257}.

\bibitem[Chorin and Tu(2009)]{chorin2009implicit}
A.~J. Chorin and X.~Tu.
\newblock Implicit sampling for particle filters.
\newblock \emph{Proceedings of the National Academy of Sciences}, 106\penalty0 (41):\penalty0 17249--17254, 2009.
\newblock \doi{10.1073/pnas.0909196106}.

\bibitem[Corenflos et~al.(2021)Corenflos, Thornton, Deligiannidis, and Doucet]{corenflos2021differentiable}
A.~Corenflos, J.~Thornton, G.~Deligiannidis, and A.~Doucet.
\newblock Differentiable particle filtering via entropy-regularized optimal transport.
\newblock In \emph{Proceedings of the 38th International Conference on Machine Learning (ICML)}, volume 139 of \emph{Proceedings of Machine Learning Research}, pages 2100--2111. PMLR, 2021.

\bibitem[Cornebise et~al.(2008)Cornebise, Moulines, and Olsson]{cornebise2008adaptive}
J.~Cornebise, {\'E}.~Moulines, and J.~Olsson.
\newblock Adaptive methods for sequential importance sampling with application to state space models.
\newblock \emph{Statistics and Computing}, 18\penalty0 (4):\penalty0 461--480, 2008.
\newblock \doi{10.1007/s11222-008-9089-4}.

\bibitem[Courtier et~al.(1994)Courtier, Th{\'e}paut, and Hollingsworth]{courtier1994strategy}
P.~Courtier, J.-N. Th{\'e}paut, and A.~Hollingsworth.
\newblock A strategy for operational implementation of {4D}-{V}ar, using an incremental approach.
\newblock \emph{Quarterly Journal of the Royal Meteorological Society}, 120\penalty0 (519):\penalty0 1367--1387, 1994.
\newblock \doi{10.1002/qj.49712051912}.

\bibitem[de~Wit and van Diepen(2007)]{DeWitVanDiepen2007CropEnKF}
A.~J.~W. de~Wit and C.~A. van Diepen.
\newblock Crop model data assimilation with the {Ensemble Kalman Filter} for improving regional crop yield forecasts.
\newblock \emph{Agricultural and Forest Meteorology}, 146\penalty0 (1--2):\penalty0 38--56, 2007.
\newblock \doi{10.1016/j.agrformet.2007.05.004}.

\bibitem[Dee et~al.(2011)Dee, Uppala, Simmons, Berrisford, Poli, Kobayashi, Andrae, Balmaseda, Balsamo, Bauer, Bechtold, Beljaars, van~de Berg, Bidlot, Bormann, Delsol, Dragani, Fuentes, Geer, Haimberger, Healy, Hersbach, H{\'o}lm, Isaksen, K{\aa}llberg, K{\"o}hler, Matricardi, McNally, Monge-Sanz, Morcrette, Park, Peubey, de~Rosnay, Tavolato, Th{\'e}paut, and Vitart]{Dee2011ERAInterim}
D.~P. Dee, S.~M. Uppala, A.~J. Simmons, P.~Berrisford, P.~Poli, S.~Kobayashi, U.~Andrae, M.~A. Balmaseda, G.~Balsamo, P.~Bauer, P.~Bechtold, A.~C.~M. Beljaars, L.~van~de Berg, J.~Bidlot, N.~Bormann, C.~Delsol, R.~Dragani, M.~Fuentes, A.~J. Geer, L.~Haimberger, S.~B. Healy, H.~Hersbach, E.~V. H{\'o}lm, L.~Isaksen, P.~K{\aa}llberg, M.~K{\"o}hler, M.~Matricardi, A.~P. McNally, B.~M. Monge-Sanz, J.-J. Morcrette, B.-K. Park, C.~Peubey, P.~de~Rosnay, C.~Tavolato, J.-N. Th{\'e}paut, and F.~Vitart.
\newblock The {ERA-Interim} reanalysis: configuration and performance of the data assimilation system.
\newblock \emph{Quarterly Journal of the Royal Meteorological Society}, 137\penalty0 (656):\penalty0 553--597, 2011.
\newblock \doi{10.1002/qj.828}.

\bibitem[Dellaert et~al.(1999)Dellaert, Fox, Burgard, and Thrun]{772544}
F.~Dellaert, D.~Fox, W.~Burgard, and S.~Thrun.
\newblock Monte carlo localization for mobile robots.
\newblock In \emph{Proceedings 1999 IEEE International Conference on Robotics and Automation (Cat. No.99CH36288C)}, volume~2, pages 1322--1328 vol.2, 1999.
\newblock \doi{10.1109/ROBOT.1999.772544}.

\bibitem[Dlamini et~al.(2023)Dlamini, Crespo, van Dam, and Kooistra]{Dlamini15164066}
L.~Dlamini, O.~Crespo, J.~van Dam, and L.~Kooistra.
\newblock A global systematic review of improving crop model estimations by assimilating remote sensing data: Implications for small-scale agricultural systems.
\newblock \emph{Remote Sensing}, 15\penalty0 (16), 2023.
\newblock ISSN 2072-4292.
\newblock \doi{10.3390/rs15164066}.
\newblock URL \url{https://www.mdpi.com/2072-4292/15/16/4066}.

\bibitem[Doucet and Johansen(2011)]{doucet2009tutorial}
A.~Doucet and A.~M. Johansen.
\newblock A tutorial on particle filtering and smoothing: Fifteen years later.
\newblock In D.~Crisan and B.~Rozovskii, editors, \emph{The Oxford Handbook of Nonlinear Filtering}, pages 656--704. Oxford University Press, Oxford, 2011.

\bibitem[Doucet et~al.(2000)Doucet, Godsill, and Andrieu]{doucet2000sequential}
A.~Doucet, S.~Godsill, and C.~Andrieu.
\newblock On sequential {M}onte {C}arlo sampling methods for {B}ayesian filtering.
\newblock \emph{Statistics and Computing}, 10\penalty0 (3):\penalty0 197--208, 2000.

\bibitem[Doucet et~al.(2001)Doucet, de~Freitas, and Gordon]{doucet2001smc}
A.~Doucet, N.~de~Freitas, and N.~Gordon.
\newblock \emph{Sequential Monte Carlo Methods in Practice}.
\newblock Springer, New York, 2001.
\newblock \doi{10.1007/978-1-4757-3437-9}.

\bibitem[Drton and Maathuis(2017)]{drton2017structure}
M.~Drton and M.~H. Maathuis.
\newblock Structure learning in graphical modeling.
\newblock \emph{Annual Review of Statistics and Its Application}, 4\penalty0 (1):\penalty0 365--393, 2017.
\newblock \doi{10.1146/annurev-statistics-060116-053803}.

\bibitem[Evensen(1994)]{evensen1994enkf}
G.~Evensen.
\newblock Sequential data assimilation with a nonlinear quasi-geostrophic model using monte carlo methods.
\newblock \emph{Journal of Geophysical Research: Oceans}, 99\penalty0 (C5):\penalty0 10143--10162, 1994.
\newblock \doi{10.1029/94JC00572}.
\newblock URL \url{https://doi.org/10.1029/94JC00572}.

\bibitem[Farchi and Bocquet(2018)]{farchi2018comparison}
A.~Farchi and M.~Bocquet.
\newblock Review article: Comparison of local particle filters and new implementations.
\newblock \emph{Nonlinear Processes in Geophysics}, 25\penalty0 (4):\penalty0 765--807, 2018.
\newblock \doi{10.5194/npg-25-765-2018}.

\bibitem[Frans et~al.(2025)Frans, Hafner, Levine, and Abbeel]{frans2024shortcut}
K.~Frans, D.~Hafner, S.~Levine, and P.~Abbeel.
\newblock One step diffusion via shortcut models.
\newblock In \emph{International Conference on Learning Representations (ICLR)}, 2025.
\newblock arXiv:2410.12557.

\bibitem[Gaspari and Cohn(1999)]{gaspari1999construction}
G.~Gaspari and S.~E. Cohn.
\newblock Construction of correlation functions in two and three dimensions.
\newblock \emph{Quarterly Journal of the Royal Meteorological Society}, 125\penalty0 (554):\penalty0 723--757, 1999.
\newblock \doi{10.1002/qj.49712555417}.

\bibitem[Geng et~al.(2025)Geng, Deng, Bai, Kolter, and He]{geng2025meanflow}
Z.~Geng, M.~Deng, X.~Bai, J.~Z. Kolter, and K.~He.
\newblock Mean flows for one-step generative modeling.
\newblock In \emph{Advances in Neural Information Processing Systems}, volume~38, 2025.
\newblock arXiv:2505.13447.

\bibitem[Gordon et~al.(1993)Gordon, Salmond, and Smith]{gordon1993novel}
N.~Gordon, D.~Salmond, and A.~Smith.
\newblock Novel approach to nonlinear/non-gaussian bayesian state estimation.
\newblock \emph{IEE Proceedings F (Radar and Signal Processing)}, 140:\penalty0 107--113, 1993.
\newblock \doi{10.1049/ip-f-2.1993.0015}.
\newblock URL \url{https://digital-library.theiet.org/doi/abs/10.1049/ip-f-2.1993.0015}.

\bibitem[Gottwald et~al.(2025)Gottwald, Liu, Marzouk, Reich, and Tong]{gottwald2025localized}
G.~A. Gottwald, S.~Liu, Y.~Marzouk, S.~Reich, and X.~T. Tong.
\newblock Localized diffusion models.
\newblock \emph{arXiv preprint arXiv:2505.04417}, 2025.
\newblock \doi{10.48550/arXiv.2505.04417}.

\bibitem[Gu et~al.(2015)Gu, Ghahramani, and Turner]{gu2015nasmc}
S.~S. Gu, Z.~Ghahramani, and R.~E. Turner.
\newblock Neural adaptive sequential {M}onte {C}arlo.
\newblock In \emph{Advances in Neural Information Processing Systems}, volume~28, 2015.

\bibitem[Hersbach et~al.(2020)Hersbach, Bell, Berrisford, Hirahara, Hor{\'a}nyi, Mu{\~n}oz-Sabater, Nicolas, Peubey, Radu, Schepers, Simmons, Soci, Abdalla, Abellan, Balsamo, Bechtold, Biavati, Bidlot, Bonavita, De~Chiara, Dahlgren, Dee, Diamantakis, Dragani, Flemming, Forbes, Fuentes, Geer, Haimberger, Healy, Hogan, H{\'o}lm, Janiskov{\'a}, Keeley, Laloyaux, Lopez, Lupu, Radnoti, de~Rosnay, Rozum, Vamborg, Villaume, and Th{\'e}paut]{Hersbach2020ERA5}
H.~Hersbach, B.~Bell, P.~Berrisford, S.~Hirahara, A.~Hor{\'a}nyi, J.~Mu{\~n}oz-Sabater, J.~Nicolas, C.~Peubey, R.~Radu, D.~Schepers, A.~Simmons, C.~Soci, S.~Abdalla, X.~Abellan, G.~Balsamo, P.~Bechtold, G.~Biavati, J.~Bidlot, M.~Bonavita, G.~De~Chiara, P.~Dahlgren, D.~Dee, M.~Diamantakis, R.~Dragani, J.~Flemming, R.~Forbes, M.~Fuentes, A.~Geer, L.~Haimberger, S.~Healy, R.~J. Hogan, E.~H{\'o}lm, M.~Janiskov{\'a}, S.~Keeley, P.~Laloyaux, P.~Lopez, C.~Lupu, G.~Radnoti, P.~de~Rosnay, I.~Rozum, F.~Vamborg, S.~Villaume, and J.-N. Th{\'e}paut.
\newblock The {ERA5} global reanalysis.
\newblock \emph{Quarterly Journal of the Royal Meteorological Society}, 146\penalty0 (730):\penalty0 1999--2049, 2020.
\newblock \doi{10.1002/qj.3803}.

\bibitem[Ho and Salimans(2021)]{ho2021classifierfree}
J.~Ho and T.~Salimans.
\newblock Classifier-free diffusion guidance.
\newblock In \emph{NeurIPS 2021 Workshop on Deep Generative Models and Downstream Applications}, 2021.
\newblock URL \url{https://openreview.net/forum?id=qw8AKxfYbI}.

\bibitem[Houtekamer and Mitchell(2001)]{ASequentialEnsembleKalmanFilterforAtmosphericDataAssimilation}
P.~L. Houtekamer and H.~L. Mitchell.
\newblock A sequential ensemble kalman filter for atmospheric data assimilation.
\newblock \emph{Monthly Weather Review}, 129\penalty0 (1):\penalty0 123 -- 137, 2001.
\newblock \doi{10.1175/1520-0493(2001)129<0123:ASEKFF>2.0.CO;2}.
\newblock URL \url{https://journals.ametsoc.org/view/journals/mwre/129/1/1520-0493_2001_129_0123_asekff_2.0.co_2.xml}.

\bibitem[Houtekamer et~al.(2014{\natexlab{a}})Houtekamer, Deng, Mitchell, Baek, and Gagnon]{HigherResolutioninanOperationalEnsembleKalmanFilter}
P.~L. Houtekamer, X.~Deng, H.~L. Mitchell, S.-J. Baek, and N.~Gagnon.
\newblock Higher resolution in an operational ensemble kalman filter.
\newblock \emph{Monthly Weather Review}, 142\penalty0 (3):\penalty0 1143 -- 1162, 2014{\natexlab{a}}.
\newblock \doi{10.1175/MWR-D-13-00138.1}.
\newblock URL \url{https://journals.ametsoc.org/view/journals/mwre/142/3/mwr-d-13-00138.1.xml}.

\bibitem[Houtekamer et~al.(2014{\natexlab{b}})Houtekamer, Deng, Mitchell, Baek, and Gagnon]{houtekamer2014higher}
P.~L. Houtekamer, X.~Deng, H.~L. Mitchell, S.-J. Baek, and N.~Gagnon.
\newblock Higher resolution in an operational ensemble {K}alman filter.
\newblock \emph{Monthly Weather Review}, 142:\penalty0 1143--1162, 2014{\natexlab{b}}.

\bibitem[Huang et~al.(2024)Huang, Gianinazzi, Yu, Dueben, and Hoefler]{huang2024diffda}
L.~Huang, L.~Gianinazzi, Y.~Yu, P.~D. Dueben, and T.~Hoefler.
\newblock Diff{DA}: a diffusion model for weather-scale data assimilation.
\newblock In \emph{Forty-first International Conference on Machine Learning}, 2024.
\newblock URL \url{https://openreview.net/forum?id=vhMq3eAB34}.

\bibitem[Hunt et~al.(2007)Hunt, Kostelich, and Szunyogh]{hunt2007efficient}
B.~R. Hunt, E.~J. Kostelich, and I.~Szunyogh.
\newblock Efficient data assimilation for spatiotemporal chaos: A local ensemble transform {K}alman filter.
\newblock \emph{Physica D: Nonlinear Phenomena}, 230\penalty0 (1--2):\penalty0 112--126, 2007.
\newblock \doi{10.1016/j.physd.2006.11.008}.

\bibitem[Hutchinson(1990)]{hutchinson1990stochastic}
M.~F. Hutchinson.
\newblock A stochastic estimator of the trace of the influence matrix for {Laplacian} smoothing splines.
\newblock \emph{Communications in Statistics - Simulation and Computation}, 19\penalty0 (2):\penalty0 433--450, 1990.
\newblock \doi{10.1080/03610919008812866}.

\bibitem[Jazwinski(1970)]{jazwinski1970stochastic}
A.~H. Jazwinski.
\newblock \emph{Stochastic Processes and Filtering Theory}, volume~64 of \emph{Mathematics in Science and Engineering}.
\newblock Academic Press, New York, 1970.

\bibitem[Jonschkowski et~al.(2018)Jonschkowski, Rastogi, and Brock]{jonschkowski2018differentiable}
R.~Jonschkowski, D.~Rastogi, and O.~Brock.
\newblock Differentiable particle filters: End-to-end learning with algorithmic priors.
\newblock In \emph{Proceedings of Robotics: Science and Systems (RSS)}, Pittsburgh, Pennsylvania, 2018.
\newblock \doi{10.15607/RSS.2018.XIV.001}.

\bibitem[Julier and Uhlmann(1997)]{julier1997ukf}
S.~J. Julier and J.~K. Uhlmann.
\newblock New extension of the {K}alman filter to nonlinear systems.
\newblock In \emph{Signal Processing, Sensor Fusion, and Target Recognition VI}, volume 3068 of \emph{Proc. SPIE}, pages 182--193, 1997.
\newblock \doi{10.1117/12.280797}.

\bibitem[Kalman(1960)]{kalman1960kf}
R.~E. Kalman.
\newblock A new approach to linear filtering and prediction problems.
\newblock \emph{Journal of Basic Engineering}, 82\penalty0 (1):\penalty0 35--45, 1960.
\newblock \doi{10.1115/1.3662552}.

\bibitem[Karkus et~al.(2018)Karkus, Hsu, and Lee]{karkus2018particle}
P.~Karkus, D.~Hsu, and W.~S. Lee.
\newblock Particle filter networks with application to visual localization.
\newblock In \emph{Proceedings of the 2nd Conference on Robot Learning (CoRL)}, volume~87 of \emph{Proceedings of Machine Learning Research}, pages 169--178. PMLR, 2018.

\bibitem[Kuramoto(1978)]{kuramoto1978}
Y.~Kuramoto.
\newblock Diffusion-induced chaos in reaction systems.
\newblock \emph{Progress of Theoretical Physics Supplement}, 64:\penalty0 346--367, 1978.
\newblock \doi{10.1143/PTPS.64.346}.

\bibitem[Lauritzen(1996)]{lauritzen1996graphical}
S.~L. Lauritzen.
\newblock \emph{Graphical Models}.
\newblock Number~17 in Oxford Statistical Science Series. Clarendon Press, Oxford, 1996.

\bibitem[Le et~al.(2018)Le, Igl, Rainforth, Jin, and Wood]{le2018aesmc}
T.~A. Le, M.~Igl, T.~Rainforth, T.~Jin, and F.~Wood.
\newblock Auto-encoding sequential {M}onte {C}arlo.
\newblock In \emph{International Conference on Learning Representations}, 2018.
\newblock URL \url{https://openreview.net/forum?id=BJ8c3f-0b}.

\bibitem[Lipman et~al.(2023)Lipman, Chen, Ben-Hamu, Nickel, and Le]{lipman2023flow}
Y.~Lipman, R.~T.~Q. Chen, H.~Ben-Hamu, M.~Nickel, and M.~Le.
\newblock Flow matching for generative modeling.
\newblock In \emph{The Eleventh International Conference on Learning Representations}, 2023.
\newblock URL \url{https://openreview.net/forum?id=PqvMRDCJT9t}.

\bibitem[Lipman et~al.(2024)Lipman, Havasi, Holderrieth, Shaul, Le, Karrer, Chen, Lopez-Paz, Ben-Hamu, and Gat]{lipman2024flowmatchingguidecode}
Y.~Lipman, M.~Havasi, P.~Holderrieth, N.~Shaul, M.~Le, B.~Karrer, R.~T.~Q. Chen, D.~Lopez-Paz, H.~Ben-Hamu, and I.~Gat.
\newblock Flow matching guide and code, 2024.
\newblock URL \url{https://arxiv.org/abs/2412.06264}.

\bibitem[Liu and Chen(1998)]{liu1998sequential}
J.~S. Liu and R.~Chen.
\newblock Sequential {Monte Carlo} methods for dynamic systems.
\newblock \emph{Journal of the American Statistical Association}, 93\penalty0 (443):\penalty0 1032--1044, 1998.
\newblock \doi{10.1080/01621459.1998.10473765}.

\bibitem[Liu et~al.(2023)Liu, Gong, and qiang liu]{liu2023flow}
X.~Liu, C.~Gong, and qiang liu.
\newblock Flow straight and fast: Learning to generate and transfer data with rectified flow.
\newblock In \emph{The Eleventh International Conference on Learning Representations}, 2023.
\newblock URL \url{https://openreview.net/forum?id=XVjTT1nw5z}.

\bibitem[Lorenc(1986{\natexlab{a}})]{Lorenc1986AnalysisMethods}
A.~C. Lorenc.
\newblock Analysis methods for numerical weather prediction.
\newblock \emph{Quarterly Journal of the Royal Meteorological Society}, 112\penalty0 (474):\penalty0 1177--1194, 1986{\natexlab{a}}.
\newblock \doi{10.1002/qj.49711247414}.

\bibitem[Lorenc(1986{\natexlab{b}})]{lorenc1986analysis}
A.~C. Lorenc.
\newblock Analysis methods for numerical weather prediction.
\newblock \emph{Quarterly Journal of the Royal Meteorological Society}, 112\penalty0 (474):\penalty0 1177--1194, 1986{\natexlab{b}}.

\bibitem[Lorenz(1963{\natexlab{a}})]{lorenz1963deterministic}
E.~N. Lorenz.
\newblock Deterministic nonperiodic flow.
\newblock \emph{Journal of the Atmospheric Sciences}, 20\penalty0 (2):\penalty0 130--141, 1963{\natexlab{a}}.
\newblock \doi{10.1175/1520-0469(1963)020<0130:DNF>2.0.CO;2}.

\bibitem[Lorenz(1963{\natexlab{b}})]{lorenz63}
E.~N. Lorenz.
\newblock Deterministic nonperiodic flow.
\newblock \emph{Journal of the Atmospheric Sciences}, 20\penalty0 (2):\penalty0 130--141, 1963{\natexlab{b}}.
\newblock \doi{10.1175/1520-0469(1963)020<0130:DNF>2.0.CO;2}.

\bibitem[Lorenz(1995)]{75462}
E.~N. Lorenz.
\newblock Predictability: A problem partly solved.
\newblock In \emph{Proceedings of Seminar on Predictability}, pages 1--18, Reading, UK, 1995. ECMWF.

\bibitem[Lorenz(1996)]{lorenz1995predictability}
E.~N. Lorenz.
\newblock Predictability: A problem partly solved.
\newblock In \emph{Proc. Seminar on Predictability}, volume~1, pages 1--18, Reading, UK, 1996. ECMWF.

\bibitem[Maddison et~al.(2017)Maddison, Lawson, Tucker, Heess, Norouzi, Mnih, Doucet, and Teh]{maddison2017filtering}
C.~J. Maddison, D.~Lawson, G.~Tucker, N.~Heess, M.~Norouzi, A.~Mnih, A.~Doucet, and Y.~W. Teh.
\newblock Filtering variational objectives.
\newblock In \emph{Advances in Neural Information Processing Systems}, volume~30, 2017.

\bibitem[Matheson and Winkler(1976)]{matheson1976scoring}
J.~E. Matheson and R.~L. Winkler.
\newblock Scoring rules for continuous probability distributions.
\newblock \emph{Management Science}, 22\penalty0 (10):\penalty0 1087--1096, 1976.
\newblock \doi{10.1287/mnsc.22.10.1087}.

\bibitem[Morzfeld et~al.(2012)Morzfeld, Tu, Atkins, and Chorin]{morzfeld2012randommap}
M.~Morzfeld, X.~Tu, E.~Atkins, and A.~J. Chorin.
\newblock A random map implementation of implicit filters.
\newblock \emph{Journal of Computational Physics}, 231\penalty0 (4):\penalty0 2049--2066, 2012.
\newblock \doi{10.1016/j.jcp.2011.11.022}.

\bibitem[Naesseth et~al.(2018)Naesseth, Linderman, Ranganath, and Blei]{naesseth2018variational}
C.~A. Naesseth, S.~W. Linderman, R.~Ranganath, and D.~M. Blei.
\newblock Variational sequential {M}onte {C}arlo.
\newblock In \emph{Proceedings of the 21st International Conference on Artificial Intelligence and Statistics (AISTATS)}, volume~84 of \emph{Proceedings of Machine Learning Research}, pages 968--977. PMLR, 2018.

\bibitem[Paige and Wood(2016)]{paige2016inference}
B.~Paige and F.~Wood.
\newblock Inference networks for sequential {M}onte {C}arlo in graphical models.
\newblock In \emph{Proceedings of the 33rd International Conference on Machine Learning}, volume~48 of \emph{Proceedings of Machine Learning Research}, pages 3040--3049. PMLR, 2016.

\bibitem[Penny and Miyoshi(2016)]{penny2016local}
S.~G. Penny and T.~Miyoshi.
\newblock A local particle filter for high-dimensional geophysical systems.
\newblock \emph{Nonlinear Processes in Geophysics}, 23\penalty0 (6):\penalty0 391--405, 2016.
\newblock \doi{10.5194/npg-23-391-2016}.

\bibitem[Perez et~al.(2018)Perez, Strub, de~Vries, Dumoulin, and Courville]{Perez_Strub_deVries_Dumoulin_Courville_2018}
E.~Perez, F.~Strub, H.~de~Vries, V.~Dumoulin, and A.~Courville.
\newblock Film: Visual reasoning with a general conditioning layer.
\newblock \emph{Proceedings of the AAAI Conference on Artificial Intelligence}, 32\penalty0 (1), Apr. 2018.
\newblock \doi{10.1609/aaai.v32i1.11671}.
\newblock URL \url{https://ojs.aaai.org/index.php/AAAI/article/view/11671}.

\bibitem[Pitt and Shephard(1999)]{pitt1999auxiliary}
M.~K. Pitt and N.~Shephard.
\newblock Filtering via simulation: Auxiliary particle filters.
\newblock \emph{Journal of the American Statistical Association}, 94\penalty0 (446):\penalty0 590--599, 1999.
\newblock \doi{10.1080/01621459.1999.10474153}.

\bibitem[Poterjoy(2016)]{poterjoy2016localized}
J.~Poterjoy.
\newblock A localized particle filter for high-dimensional nonlinear systems.
\newblock \emph{Monthly Weather Review}, 144\penalty0 (1):\penalty0 59--76, 2016.
\newblock \doi{10.1175/MWR-D-15-0163.1}.

\bibitem[Rabier et~al.(2000)Rabier, J{\"a}rvinen, Klinker, Mahfouf, and Simmons]{Rabier2000ECMWF4DVar}
F.~Rabier, H.~J{\"a}rvinen, E.~Klinker, J.-F. Mahfouf, and A.~Simmons.
\newblock The {ECMWF} operational implementation of four-dimensional variational assimilation. {I}: Experimental results with simplified physics.
\newblock \emph{Quarterly Journal of the Royal Meteorological Society}, 126\penalty0 (564):\penalty0 1143--1170, 2000.
\newblock \doi{10.1256/smsqj.56414}.

\bibitem[Rebeschini and van Handel(2015)]{rebeschini2015local}
P.~Rebeschini and R.~van Handel.
\newblock Can local particle filters beat the curse of dimensionality?
\newblock \emph{The Annals of Applied Probability}, 25\penalty0 (5):\penalty0 2809--2866, 2015.
\newblock \doi{10.1214/14-AAP1061}.

\bibitem[Rozet and Louppe(2023)]{rozet2023scorebased}
F.~Rozet and G.~Louppe.
\newblock Score-based data assimilation.
\newblock In \emph{Thirty-seventh Conference on Neural Information Processing Systems}, 2023.
\newblock URL \url{https://openreview.net/forum?id=VUvLSnMZdX}.

\bibitem[Runge et~al.(2019)Runge, Bathiany, Bollt, Camps-Valls, Coumou, Deyle, Glymour, Kretschmer, Mahecha, Mu{\~n}oz-Mar{\'\i}, van Nes, Peters, Quax, Reichstein, Scheffer, Sch{\"o}lkopf, Spirtes, Sugihara, Sun, Zhang, and Zscheischler]{runge2019inferring}
J.~Runge, S.~Bathiany, E.~Bollt, G.~Camps-Valls, D.~Coumou, E.~Deyle, C.~Glymour, M.~Kretschmer, M.~D. Mahecha, J.~Mu{\~n}oz-Mar{\'\i}, E.~H. van Nes, J.~Peters, R.~Quax, M.~Reichstein, M.~Scheffer, B.~Sch{\"o}lkopf, P.~Spirtes, G.~Sugihara, J.~Sun, K.~Zhang, and J.~Zscheischler.
\newblock Inferring causation from time series in {E}arth system sciences.
\newblock \emph{Nature Communications}, 10\penalty0 (1):\penalty0 2553, 2019.
\newblock \doi{10.1038/s41467-019-10105-3}.

\bibitem[Saharia et~al.(2022)Saharia, Chan, Saxena, Li, Whang, Denton, Ghasemipour, Gontijo~Lopes, Karagol~Ayan, Salimans, Ho, Fleet, and Norouzi]{saharia2022photorealistic}
C.~Saharia, W.~Chan, S.~Saxena, L.~Li, J.~Whang, E.~L. Denton, S.~K.~S. Ghasemipour, R.~Gontijo~Lopes, B.~Karagol~Ayan, T.~Salimans, J.~Ho, D.~J. Fleet, and M.~Norouzi.
\newblock Photorealistic text-to-image diffusion models with deep language understanding.
\newblock In \emph{Advances in Neural Information Processing Systems}, volume~35, pages 36479--36494, 2022.

\bibitem[S{\"a}rkk{\"a}(2013)]{sarkka2013bayesian}
S.~S{\"a}rkk{\"a}.
\newblock \emph{Bayesian Filtering and Smoothing}.
\newblock Institute of Mathematical Statistics Textbooks. Cambridge University Press, 2013.
\newblock \doi{10.1017/CBO9781139344203}.

\bibitem[Savary et~al.(2026)Savary, Rozet, and Louppe]{savary2026trainingfree}
T.~Savary, F.~Rozet, and G.~Louppe.
\newblock Training-free bayesian filtering with generative emulators.
\newblock In \emph{Proceedings of the 43rd International Conference on Machine Learning (ICML)}, 2026.
\newblock URL \url{https://arxiv.org/abs/2605.20028}.

\bibitem[Schraff et~al.(2016)Schraff, Reich, Rhodin, Schomburg, Stephan, Peri{\'a}{\~n}ez, and Potthast]{schraff2016kilometre}
C.~Schraff, H.~Reich, A.~Rhodin, A.~Schomburg, K.~Stephan, A.~Peri{\'a}{\~n}ez, and R.~Potthast.
\newblock Kilometre-scale ensemble data assimilation for the {COSMO} model ({KENDA}).
\newblock \emph{Quarterly Journal of the Royal Meteorological Society}, 142:\penalty0 1453--1472, 2016.

\bibitem[Si and Chen(2025)]{si2025latentensf}
P.~Si and P.~Chen.
\newblock Latent-en{SF}: A latent ensemble score filter for high-dimensional data assimilation with sparse observation data.
\newblock In \emph{The Thirteenth International Conference on Learning Representations}, 2025.
\newblock URL \url{https://openreview.net/forum?id=urcEYsZOBz}.

\bibitem[Sivashinsky(1977)]{sivashinsky1977}
G.~I. Sivashinsky.
\newblock Nonlinear analysis of hydrodynamic instability in laminar flames---{I}. {D}erivation of basic equations.
\newblock \emph{Acta Astronautica}, 4\penalty0 (11--12):\penalty0 1177--1206, 1977.
\newblock \doi{10.1016/0094-5765(77)90096-0}.

\bibitem[Snyder et~al.(2008{\natexlab{a}})Snyder, Bengtsson, Bickel, and Anderson]{ObstaclestoHighDimensionalParticleFiltering}
C.~Snyder, T.~Bengtsson, P.~Bickel, and J.~Anderson.
\newblock Obstacles to high-dimensional particle filtering.
\newblock \emph{Monthly Weather Review}, 136\penalty0 (12):\penalty0 4629 -- 4640, 2008{\natexlab{a}}.
\newblock \doi{10.1175/2008MWR2529.1}.
\newblock URL \url{https://journals.ametsoc.org/view/journals/mwre/136/12/2008mwr2529.1.xml}.

\bibitem[Snyder et~al.(2008{\natexlab{b}})Snyder, Bengtsson, Bickel, and Anderson]{snyder2008obstacles}
C.~Snyder, T.~Bengtsson, P.~Bickel, and J.~Anderson.
\newblock Obstacles to high-dimensional particle filtering.
\newblock \emph{Monthly Weather Review}, 136\penalty0 (12):\penalty0 4629--4640, 2008{\natexlab{b}}.
\newblock \doi{10.1175/2008MWR2529.1}.

\bibitem[Snyder et~al.(2015)Snyder, Bengtsson, and Morzfeld]{snyder2015performance}
C.~Snyder, T.~Bengtsson, and M.~Morzfeld.
\newblock Performance bounds for particle filters using the optimal proposal.
\newblock \emph{Monthly Weather Review}, 143\penalty0 (11):\penalty0 4750--4761, 2015.
\newblock \doi{10.1175/MWR-D-15-0144.1}.

\bibitem[Song et~al.(2021)Song, Sohl-Dickstein, Kingma, Kumar, Ermon, and Poole]{song2021scorebased}
Y.~Song, J.~Sohl-Dickstein, D.~P. Kingma, A.~Kumar, S.~Ermon, and B.~Poole.
\newblock Score-based generative modeling through stochastic differential equations.
\newblock In \emph{International Conference on Learning Representations}, 2021.
\newblock URL \url{https://openreview.net/forum?id=PxTIG12RRHS}.

\bibitem[Song et~al.(2023)Song, Dhariwal, Chen, and Sutskever]{song2023consistency}
Y.~Song, P.~Dhariwal, M.~Chen, and I.~Sutskever.
\newblock Consistency models.
\newblock In \emph{Proceedings of the 40th International Conference on Machine Learning (ICML)}, volume 202 of \emph{Proceedings of Machine Learning Research}, pages 32211--32252. PMLR, 2023.

\bibitem[Tang et~al.(2026)Tang, Bausback, Bao, Zhang, and Huynh]{tang2026score}
J.~Tang, R.~Bausback, F.~Bao, G.~Zhang, and P.-T. Huynh.
\newblock A score filter enhanced data assimilation framework for data-driven dynamical systems.
\newblock \emph{arXiv preprint arXiv:2603.14863}, Mar. 2026.
\newblock URL \url{https://arxiv.org/abs/2603.14863}.

\bibitem[Transue et~al.(2025)Transue, Chen, Takao, and Wang]{transue_flow_2025}
T.~Transue, B.~Chen, S.~Takao, and B.~Wang.
\newblock Flow matching for efficient and scalable data assimilation.
\newblock \emph{arXiv preprint arXiv:2508.13313}, 2025.
\newblock \doi{10.48550/arXiv.2508.13313}.
\newblock URL \url{https://arxiv.org/abs/2508.13313}.

\bibitem[van Leeuwen(2010)]{vanleeuwen2010ewpf}
P.~J. van Leeuwen.
\newblock Nonlinear data assimilation in geosciences: an extremely efficient particle filter.
\newblock \emph{Quarterly Journal of the Royal Meteorological Society}, 136\penalty0 (653):\penalty0 1991--1999, 2010.
\newblock \doi{10.1002/qj.699}.

\bibitem[van Leeuwen et~al.(2019)van Leeuwen, K\"unsch, Nerger, Potthast, and Reich]{vanleeuwen2019review}
P.~J. van Leeuwen, H.~R. K\"unsch, L.~Nerger, R.~Potthast, and S.~Reich.
\newblock Particle filters for high-dimensional geoscience applications: A review.
\newblock \emph{Quarterly Journal of the Royal Meteorological Society}, 145\penalty0 (723):\penalty0 2335--2365, 2019.
\newblock \doi{10.1002/qj.3551}.

\bibitem[Xiao et~al.(2024)Xiao, Si, and Chen]{xiao2024ldensf}
P.~Xiao, P.~Si, and P.~Chen.
\newblock {LD}-{E}n{SF}: Synergizing latent dynamics with ensemble score filters for fast data assimilation with sparse observations.
\newblock \emph{arXiv preprint arXiv:2411.19305}, 2024.
\newblock \doi{10.48550/arXiv.2411.19305}.

\end{thebibliography}

\appendix

\section{Experimental Setup}
\label{app:common}

\subsection{Data split}
For each system we generate 2048 simulated trajectories and split 80/10/10 into
train/validation/test. A subset of the test split is used to report results due
to compute constraints; exact subset sizes are listed per experiment.

\subsection{FPPF architecture and hyperparameters}
\label{app:architectures}

\paragraph{Backbone.}
For architectural parity, the same backbone with matched parameter count is
used as the velocity field in FPPF, the drift network in FlowDAS, and the
conditional network in NASMC and InfNN, with only the output head differing
across methods. Across all methods we use one of two velocity / feature backbones, picked
by spatial structure of the system. \textbf{(i) For the periodic systems
(L96 and KS)} the backbone is a 1D ResNet with circular padding and
adaptive layer normalization (AdaLN) for flow-time conditioning. Each
residual block applies $\mathrm{AdaLN}(\cdot,s)\!\to\!\mathrm{SiLU}\!\to\!
\mathrm{Conv1d}_\mathrm{circ}\!\to\!\mathrm{AdaLN}(\cdot,s)\!\to\!
\mathrm{SiLU}\!\to\!\mathrm{Conv1d}_\mathrm{circ}$ with kernel size~5
and a residual skip; AdaLN regresses per-channel $(\gamma, \beta)$ from
the flow-time embedding (zero-initialized so the block is initially the
identity). The output projection ($1{\times}1$ conv to one channel) is
also zero-initialized. \textbf{(ii) For the low-dimensional system (L63)}
the backbone is a flat MLP with SiLU activations and concatenation
conditioning. The same body is used as the velocity field in FPPF, the
drift in FlowDAS, and the conditional density backbone in NASMC and
InfNN; the methods differ only in the output head (vector velocity vs.\
mixture-of-Gaussians vs.\ RNADE).

\paragraph{Conditioning.}
We apply two training-only conditioning regularizers. {(1)}
Similar to classifier-free conditioning dropout with probability $p_\text{cond}=0.1$
the  conditioning $\bo_t$ is zeroed before being fed
to the backbone, so the network sees both conditional and unconditional
batches with respect to the observation. {(2) Previous-state corruption schedule:} during
training only, with annealed probability $p(t)=\max(p_\text{min},\,
p_0\,(1-t/T))$ we corrupt $\bx_{t-1}$ by zeroing a random subset of
its coordinates of size $\lceil r_\text{mask}\,d_x\rceil$ (no additive
Gaussian noise). We use $p_0=0.3$, $p_\text{min}=0.05$, $r_\text{mask}=0.4$, with $T$ set to the total number of training steps. This forces the
conditional network to use $\bo_t$ rather than over-relying on the
clean $\bx_{t-1}$ on systems whose forward dynamics is highly
predictable. The corruption is disabled at inference.

\paragraph{Training}
All models are trained for a minimum of 80 epochs, with an early-stopping patience set to 50 epochs.

\paragraph{L-FPPF patch network.}
The localized velocity field $u_\phi$ is a small 1D ResNet operating on
spatial patches of size $W = 2r+1$. We use radius $r = 4$ (so $W = 9$),
$C = 64$ channels, 2 residual blocks of kernel size 3, and a 64-dim
flow-time embedding fed via AdaLN. Inputs are stacked as a 4-channel
window $[\bz_{[j-r:j+r]},\,\bx_{t-1,[j-r:j+r]},\,\bo_{t,[j-r:j+r]}^{\text{full}},
\,m_{t,[j-r:j+r]}]$.

\subsection{Baseline Implementation Details}
\label{app:baselines}

\paragraph{Auxiliary particle filter (APF).}
We use the classical Pitt--Shephard \citep{pitt1999auxiliary}
auxiliary-particle filter with the prior proposal $q = p(\bx_t \mid
\bx_{t-1})$ and a point look-ahead. Concretely, the auxiliary
adjustment multiplier is
$m_t^{(i)} \;=\; g\!\left(\bo_t \,\Big|\, \bm{\mu}_t^{(i)}\right)$,
where $\bm{\mu}_t^{(i)} = \mathbb{E}\!\left[\bx_t \mid \bx_{t-1}^{(i)}\right]$
is the deterministic transition mean (one $\Delta t$ Euler step of the
true dynamics, no process noise). With $q = p$, the APF incremental
weight collapses to $\log w_t^{(i)} = \log g(\bo_t \mid \bx_t^{(i)}) -
\log m_t^{(k_i)}$, where $k_i$ is the ancestor selected in the
first-stage resample drawn from $\lambda^{(i)} \propto w_{t-1}^{(i)}
m_t^{(i)}$. Weights
are reset to the second-stage incremental weight after each step, and a
final $\textsc{ess}<0.33\,N$-triggered systematic resample is applied to
keep the cloud healthy.

\paragraph{Ensemble Kalman filter (EnKF).}
We use EnKF with multiplicative covariance inflation. The inflation
factor is tuned over $\lambda \in \{0.8, 0.9, 1.0, 1.1\}$ on the validation set and reported at the best-performing $\lambda$ (lowest RMSE), ensemble size $N=50$.

\paragraph{LETKF.}
Our LETKF \citep{hunt2007efficient} uses a per-coordinate local
analysis on a 1-D periodic spatial domain. For each state index $j$,
observations are tapered by a Gaspari--Cohn polynomial
$\rho(d / r)$ with periodic distance $d$ (modulo $d_x$) and
localization radius $r$, the local symmetric square-root analysis is
solved in ensemble space, and the resulting analysis ensemble member
$\bx^{(i)}_j$ is assembled by stacking centre-coordinate analyses
We tune inflation factor on a held-out validation
subset over $\lambda \in \{0.8,0.9,\ldots,1.8\}$ and
set $r$ to match other localized baselines.
For the L96-25 in-distribution and high-dim ($d_x \in \{50, 100, 500,
1000\}$) cells we use $r=4$ (matched to the L-FPPF localization
radius of $r=4$ patches), $\lambda=1.0$, ensemble size $N=50$.

\paragraph{LBPF.}
The localized BPF is
the block-localized particle filter of \citet{farchi2018comparison}:
the spatial domain is partitioned into contiguous blocks and the
analysis weights, ancestor resampling, and patchwork particle
reassembly are computed independently within each block on tapered
local likelihoods. We use block size 1 (per-coordinate
analysis), with localization radius $r=4$ (matched to L-FPPF), and systematic
local resampling with the identity-preserving (adjustment-minimizing)
ancestor-permutation tie-break of \citet{farchi2018comparison}. No
post-regularization jitter is applied.

\paragraph{FlowDAS.}
For all FlowDAS experiments \citep{chen2025flowdas} we use the official
implementation. The guidance scale is set to 0.01 (selected via ablation),
$J=21$ Monte Carlo sampling times, learning rate $3\times10^{-4}$ with early
stopping. Inference uses 50 first-order Euler steps. FlowDAS does not produce
an ensemble by default; for CRPS we draw 20 independent rollouts per
evaluation trajectory and compute CRPS from this empirical predictive
distribution.

\paragraph{NASMC.}
We follow \citet{gu2015nasmc} and report two variants: \texttt{NASMC$_{K{=}1}$}, the
single-Gaussian baseline of \citeauthor{gu2015nasmc}, and
\texttt{NASMC$_{K{=}3}$}, the ``-MD-'' mixture-density variant with
$K=3$ components (paper default). We do \emph{not} use the paper's LSTM
history conditioning; conditioning is Markovian on $(\bx_{t-1},
\bo_t)$, matching the FPPF / InfNN comparison. At evaluation, the trained $q_\phi$ is
plugged into our standard BPF and inherits all per-cell PF settings
($N$, ESS threshold, $\sigma_x$, $\sigma_y$, batch size) from the
corresponding FPPF cell.

\paragraph{InfNN.}
We use \texttt{InfNN-RNADE} variant
from the paper: coordinates
are factorized in their natural index order
$\prod_{j=1}^{d_x} q_\phi(x_t^{(j)} \mid x_t^{(<j)}, \bx_{t-1},
\bo_t)$, with each factor a $K=3$-component diagonal Gaussian
mixture produced by a $2$-layer MLP of hidden size $128$
(\textsc{rnade} head, mirroring the original paper). At evaluation, the trained $q_\phi$ is
plugged into our standard BPF and inherits all per-cell PF settings
($N$, ESS threshold, $\sigma_x$, $\sigma_y$, batch size) from the
corresponding FPPF cell.

\subsection{Inference time proposal cost}
\label{app:baselines}
For all experiments we use $N_s = N_\ell = 32$ Euler integration steps for
both forward sampling and reverse likelihood evaluation, on a uniform
time grid for sampling and a $\gamma{=}2$ non-uniform power-back grid for
likelihood, where 	
$s_k = 1 - (1-u_k)^{\gamma}$, $\gamma > 1$. This results in smaller step sizes closer to the target distribution. Empirically, we find that the non-uniform grid allowed us to decrease the number of integration steps from 100 to 32 with minimal performance degradation, compared to a uniform grid.
A natural direction to reduce the inference-time cost of FPPF, which currently
runs $\sim$32-step Euler integration twice per particle per step (once for
sampling, once for likelihood evaluation), is to replace the rectified-flow
teacher with a few-step student. Recent work on consistency models
\citep{song2023consistency}, MeanFlow \citep{geng2025meanflow}, and Shortcut
models \citep{frans2024shortcut} produces high-quality samples in one or a
handful of NFEs, and the F2D2 framework \citep{ai2026joint} extends this idea
to fast \emph{divergence} estimation, opening a path to also accelerate the
likelihood evaluation step rather than only sampling. In our preliminary
experiments, however, we found that the resulting few-step proposals were of
substantially lower quality than the teacher: students tended to collapse
toward the conditional mean, eroding the observation-informed structure that
makes the proposal useful in the first place, and the residual bias in the
distilled divergence degraded ESS over long horizons. We therefore report the
32-step teacher throughout the main paper. Bridging the gap between
distillation quality and proposal informativeness is an interesting direction
for reducing the compute cost of FPPF.

\section{Linear-Gaussian System Details}
\label{app:lg_sanity}

\paragraph{System.}
We use $d_x{=}8$ with $\mathbf{x}_t = \mathbf{A} \mathbf{x}_{t-1} + \mathbf{w}_t$,
$\mathbf{o}_t = \mathbf{H} \mathbf{x}_t + \mathbf{v}_t$, where
$\mathbf{w}_t \sim \mathcal{N}(\mathbf{0}, \mathbf{Q})$,
$\mathbf{v}_t \sim \mathcal{N}(\mathbf{0}, \mathbf{R})$, and
$\mathbf{x}_0 \sim \mathcal{N}(\mathbf{0}, \mathbf{I})$.
Writing $\mathbf{S}$ for the cyclic shift and
$\mathbf{C}(\alpha)_{ij} = \alpha^{|i-j|}$ for a Toeplitz correlation, we set
$\mathbf{A} = 0.92\,\mathbf{I} + 0.05\,\mathbf{S} + 0.02\,\mathbf{S}^{\!\top}$,
$\mathbf{H} = \mathbf{I} + 0.25\,\mathbf{S} - 0.15\,\mathbf{S}^{\!\top}$,
$\mathbf{Q} = 0.35^2(0.7\,\mathbf{I} + 0.3\,\mathbf{C}(0.5))$,
$\mathbf{R} = 0.25^2(0.6\,\mathbf{I} + 0.4\,\mathbf{C}(0.7))$.
$\mathbf{A}$ is stable ($\rho(\mathbf{A}){<}1$), $\mathbf{H}$ mixes neighbours
asymmetrically, and both noises are spatially correlated. All states are
observed.

\paragraph{Dataset.}
$N_\text{traj}{=}1024$ trajectories of length $T{=}200$ after a $100$-step
burn-in, split $80/10/10$ by trajectory into train / val / test.

\paragraph{Optimal proposal.}
For this system $q^\star(\mathbf{x}_t \mid \mathbf{x}_{t-1}, \mathbf{o}_t)$
is Gaussian with
\[
\mathbf{K} = \mathbf{Q} \mathbf{H}^{\!\top}(\mathbf{H} \mathbf{Q} \mathbf{H}^{\!\top} + \mathbf{R})^{-1}, \quad
\boldsymbol{\mu}^\star = \mathbf{A} \mathbf{x}_{t-1} + \mathbf{K}(\mathbf{o}_t - \mathbf{H} \mathbf{A} \mathbf{x}_{t-1}), \quad
\boldsymbol{\Sigma}^\star = (\mathbf{I} - \mathbf{K} \mathbf{H})\mathbf{Q}.
\]
$\boldsymbol{\Sigma}^\star$ is independent of $(\mathbf{x}_{t-1}, \mathbf{o}_t)$,
so all conditioning enters through $\boldsymbol{\mu}^\star$. This is the
target the learned proposal $q_\phi$ must match.

\paragraph{Bootstrap reference.}
The dynamics prior $q_b(\mathbf{x}_t \mid \mathbf{x}_{t-1}) = \mathcal{N}(\mathbf{A} \mathbf{x}_{t-1}, \mathbf{Q})$
ignores $\mathbf{o}_t$ and serves as the floor a useful learned proposal must beat.

\section{Full Lorenz-96 Results}
\label{app:lorenz96_full}
Complete results are in Table~\ref{tab:lorenz96_arctan_s02} and Table~\ref{tab:lorenz96_quad_capped_s02}.
\begin{table*}[!t]
\vskip 0.1in
\centering
\small
\begin{tabular}{@{}lcccccc@{}}
\toprule
& \multicolumn{6}{c}{State Dimension $d_x$} \\
\cmidrule(l){2-7}
Method & 5 & 10 & 15 & 20 & 25 & 50 \\
\midrule
\multicolumn{7}{l}{\textit{RMSE} $\downarrow$} \\
\midrule
BPF         & 0.303 {\scriptsize $\pm$ 0.030} & 0.357 {\scriptsize $\pm$ 0.036} & 0.364 {\scriptsize $\pm$ 0.029} & 0.419 {\scriptsize $\pm$ 0.038} & 0.480 {\scriptsize $\pm$ 0.046} & 0.743 {\scriptsize $\pm$ 0.122} \\
APF         & 0.304 {\scriptsize $\pm$ 0.036} & 0.353 {\scriptsize $\pm$ 0.033} & \uline{0.356 {\scriptsize $\pm$ 0.029}} & \uline{0.401 {\scriptsize $\pm$ 0.037}} & 0.450 {\scriptsize $\pm$ 0.039} & \uline{0.673 {\scriptsize $\pm$ 0.099}} \\
EnKF        & \uline{0.287 {\scriptsize $\pm$ 0.050}} & \uline{0.326 {\scriptsize $\pm$ 0.056}} & 0.365 {\scriptsize $\pm$ 0.056} & 0.416 {\scriptsize $\pm$ 0.069} & \uline{0.448 {\scriptsize $\pm$ 0.074}} & 0.724 {\scriptsize $\pm$ 0.151} \\
NASMC-Gauss & 0.430 {\scriptsize $\pm$ 0.068} & 0.529 {\scriptsize $\pm$ 0.084} & 0.573 {\scriptsize $\pm$ 0.101} & 0.617 {\scriptsize $\pm$ 0.146} & 0.653 {\scriptsize $\pm$ 0.145} & 0.762 {\scriptsize $\pm$ 0.167} \\
NASMC-MD    & 0.471 {\scriptsize $\pm$ 0.071} & 0.533 {\scriptsize $\pm$ 0.100} & 0.577 {\scriptsize $\pm$ 0.108} & 0.614 {\scriptsize $\pm$ 0.102} & 0.659 {\scriptsize $\pm$ 0.128} & 0.751 {\scriptsize $\pm$ 0.143} \\
InfNN       & 0.524 {\scriptsize $\pm$ 0.192} & 1.000 {\scriptsize $\pm$ 0.411} & 1.363 {\scriptsize $\pm$ 0.316} & 2.117 {\scriptsize $\pm$ 0.415} & 2.467 {\scriptsize $\pm$ 0.392} & 3.396 {\scriptsize $\pm$ 0.231} \\
FPPF        & \textbf{0.189 {\scriptsize $\pm$ 0.027}} & \textbf{0.287 {\scriptsize $\pm$ 0.070}} & \textbf{0.217 {\scriptsize $\pm$ 0.043}} & \textbf{0.238 {\scriptsize $\pm$ 0.037}} & \textbf{0.235 {\scriptsize $\pm$ 0.032}} & \textbf{0.323 {\scriptsize $\pm$ 0.114}} \\
FlowDAS     & 0.438 {\scriptsize $\pm$ 0.191} & 0.562 {\scriptsize $\pm$ 0.366} & 0.505 {\scriptsize $\pm$ 0.293} & 0.735 {\scriptsize $\pm$ 0.282} & 0.603 {\scriptsize $\pm$ 0.295} & 0.970 {\scriptsize $\pm$ 0.313} \\
\midrule
\multicolumn{7}{l}{\textit{CRPS} $\downarrow$} \\
\midrule
BPF         & 0.195 {\scriptsize $\pm$ 0.013} & 0.212 {\scriptsize $\pm$ 0.018} & 0.212 {\scriptsize $\pm$ 0.012} & 0.233 {\scriptsize $\pm$ 0.018} & 0.262 {\scriptsize $\pm$ 0.023} & 0.439 {\scriptsize $\pm$ 0.078} \\
APF         & 0.195 {\scriptsize $\pm$ 0.015} & 0.210 {\scriptsize $\pm$ 0.016} & 0.209 {\scriptsize $\pm$ 0.012} & \uline{0.225 {\scriptsize $\pm$ 0.016}} & \uline{0.246 {\scriptsize $\pm$ 0.018}} & \uline{0.381 {\scriptsize $\pm$ 0.063}} \\
EnKF        & \uline{0.174 {\scriptsize $\pm$ 0.024}} & \uline{0.186 {\scriptsize $\pm$ 0.026}} & \uline{0.203 {\scriptsize $\pm$ 0.029}} & 0.229 {\scriptsize $\pm$ 0.037} & 0.246 {\scriptsize $\pm$ 0.041} & 0.422 {\scriptsize $\pm$ 0.094} \\
NASMC-Gauss & 0.263 {\scriptsize $\pm$ 0.041} & 0.332 {\scriptsize $\pm$ 0.058} & 0.358 {\scriptsize $\pm$ 0.067} & 0.393 {\scriptsize $\pm$ 0.092} & 0.423 {\scriptsize $\pm$ 0.095} & 0.501 {\scriptsize $\pm$ 0.097} \\
NASMC-MD    & 0.292 {\scriptsize $\pm$ 0.044} & 0.336 {\scriptsize $\pm$ 0.067} & 0.360 {\scriptsize $\pm$ 0.069} & 0.391 {\scriptsize $\pm$ 0.067} & 0.423 {\scriptsize $\pm$ 0.081} & 0.493 {\scriptsize $\pm$ 0.083} \\
InfNN       & 0.325 {\scriptsize $\pm$ 0.142} & 0.662 {\scriptsize $\pm$ 0.301} & 0.908 {\scriptsize $\pm$ 0.224} & 1.448 {\scriptsize $\pm$ 0.283} & 1.711 {\scriptsize $\pm$ 0.264} & 2.450 {\scriptsize $\pm$ 0.176} \\
FPPF        & \textbf{0.115 {\scriptsize $\pm$ 0.013}} & \textbf{0.165 {\scriptsize $\pm$ 0.044}} & \textbf{0.122 {\scriptsize $\pm$ 0.022}} & \textbf{0.132 {\scriptsize $\pm$ 0.018}} & \textbf{0.132 {\scriptsize $\pm$ 0.017}} & \textbf{0.183 {\scriptsize $\pm$ 0.056}} \\
FlowDAS     & 0.281 {\scriptsize $\pm$ 0.109} & 0.330 {\scriptsize $\pm$ 0.213} & 0.277 {\scriptsize $\pm$ 0.162} & 0.411 {\scriptsize $\pm$ 0.163} & 0.316 {\scriptsize $\pm$ 0.152} & 0.530 {\scriptsize $\pm$ 0.177} \\
\bottomrule
\end{tabular}
\vspace{0.2in}
\caption{Lorenz-96 results on the arctan observation operator at observation noise $\sigma_y = 0.2$ across state dimensions. Best results in \textbf{bold}; second best underlined.}
\label{tab:lorenz96_arctan_s02}
\vskip -0.1in
\end{table*}

\begin{table*}[!t]
\vskip 0.1in
\centering
\small
\begin{tabular}{@{}lcccccc@{}}
\toprule
& \multicolumn{6}{c}{State Dimension $d_x$} \\
\cmidrule(l){2-7}
Method & 5 & 10 & 15 & 20 & 25 & 50 \\
\midrule
\multicolumn{7}{l}{\textit{RMSE} $\downarrow$} \\
\midrule
BPF         & \uline{0.279 {\scriptsize $\pm$ 0.034}} & \uline{0.411 {\scriptsize $\pm$ 0.070}} & \uline{0.404 {\scriptsize $\pm$ 0.051}} & \uline{0.591 {\scriptsize $\pm$ 0.318}} & 0.803 {\scriptsize $\pm$ 0.474} & 1.942 {\scriptsize $\pm$ 0.552} \\
APF         & 0.288 {\scriptsize $\pm$ 0.032} & 0.416 {\scriptsize $\pm$ 0.061} & 0.429 {\scriptsize $\pm$ 0.071} & 0.606 {\scriptsize $\pm$ 0.367} & 0.772 {\scriptsize $\pm$ 0.471} & 1.878 {\scriptsize $\pm$ 0.560} \\
EnKF        & 0.757 {\scriptsize $\pm$ 0.976} & 1.661 {\scriptsize $\pm$ 1.580} & 2.621 {\scriptsize $\pm$ 1.543} & 3.157 {\scriptsize $\pm$ 1.319} & 3.474 {\scriptsize $\pm$ 1.149} & 4.353 {\scriptsize $\pm$ 0.531} \\
NASMC-Gauss & 0.556 {\scriptsize $\pm$ 0.332} & 0.957 {\scriptsize $\pm$ 0.690} & 1.060 {\scriptsize $\pm$ 0.650} & 1.248 {\scriptsize $\pm$ 0.699} & 1.502 {\scriptsize $\pm$ 0.707} & 1.425 {\scriptsize $\pm$ 0.486} \\
NASMC-MD    & 0.639 {\scriptsize $\pm$ 0.419} & 0.857 {\scriptsize $\pm$ 0.616} & 1.050 {\scriptsize $\pm$ 0.574} & 1.416 {\scriptsize $\pm$ 0.796} & 1.184 {\scriptsize $\pm$ 0.628} & 1.605 {\scriptsize $\pm$ 0.514} \\
InfNN       & 0.962 {\scriptsize $\pm$ 1.086} & 3.282 {\scriptsize $\pm$ 0.914} & 3.599 {\scriptsize $\pm$ 0.624} & 4.213 {\scriptsize $\pm$ 0.425} & 4.314 {\scriptsize $\pm$ 0.424} & 4.726 {\scriptsize $\pm$ 0.262} \\
FPPF        & \textbf{0.201 {\scriptsize $\pm$ 0.032}} & \textbf{0.321 {\scriptsize $\pm$ 0.125}} & \textbf{0.237 {\scriptsize $\pm$ 0.059}} & \textbf{0.382 {\scriptsize $\pm$ 0.310}} & \textbf{0.254 {\scriptsize $\pm$ 0.068}} & \textbf{0.519 {\scriptsize $\pm$ 0.298}} \\
FlowDAS     & 0.354 {\scriptsize $\pm$ 0.132} & 0.560 {\scriptsize $\pm$ 0.387} & 0.510 {\scriptsize $\pm$ 0.283} & 0.740 {\scriptsize $\pm$ 0.288} & \uline{0.629 {\scriptsize $\pm$ 0.320}} & \uline{0.945 {\scriptsize $\pm$ 0.321}} \\
\midrule
\multicolumn{7}{l}{\textit{CRPS} $\downarrow$} \\
\midrule
BPF         & \uline{0.205 {\scriptsize $\pm$ 0.013}} & \uline{0.250 {\scriptsize $\pm$ 0.036}} & \uline{0.242 {\scriptsize $\pm$ 0.023}} & \uline{0.332 {\scriptsize $\pm$ 0.176}} & 0.456 {\scriptsize $\pm$ 0.307} & 1.213 {\scriptsize $\pm$ 0.396} \\
APF         & 0.207 {\scriptsize $\pm$ 0.013} & 0.250 {\scriptsize $\pm$ 0.030} & 0.253 {\scriptsize $\pm$ 0.033} & 0.342 {\scriptsize $\pm$ 0.226} & 0.431 {\scriptsize $\pm$ 0.302} & 1.143 {\scriptsize $\pm$ 0.385} \\
EnKF        & 0.520 {\scriptsize $\pm$ 0.772} & 1.145 {\scriptsize $\pm$ 1.203} & 1.761 {\scriptsize $\pm$ 1.157} & 2.102 {\scriptsize $\pm$ 1.022} & 2.368 {\scriptsize $\pm$ 0.941} & 3.102 {\scriptsize $\pm$ 0.491} \\
NASMC-Gauss & 0.358 {\scriptsize $\pm$ 0.263} & 0.657 {\scriptsize $\pm$ 0.539} & 0.703 {\scriptsize $\pm$ 0.456} & 0.823 {\scriptsize $\pm$ 0.482} & 0.961 {\scriptsize $\pm$ 0.487} & 0.901 {\scriptsize $\pm$ 0.311} \\
NASMC-MD    & 0.423 {\scriptsize $\pm$ 0.335} & 0.581 {\scriptsize $\pm$ 0.472} & 0.697 {\scriptsize $\pm$ 0.410} & 0.920 {\scriptsize $\pm$ 0.542} & 0.773 {\scriptsize $\pm$ 0.421} & 0.995 {\scriptsize $\pm$ 0.333} \\
InfNN       & 0.684 {\scriptsize $\pm$ 0.915} & 2.441 {\scriptsize $\pm$ 0.747} & 2.606 {\scriptsize $\pm$ 0.538} & 3.149 {\scriptsize $\pm$ 0.381} & 3.227 {\scriptsize $\pm$ 0.401} & 3.651 {\scriptsize $\pm$ 0.244} \\
FPPF        & \textbf{0.125 {\scriptsize $\pm$ 0.017}} & \textbf{0.191 {\scriptsize $\pm$ 0.083}} & \textbf{0.134 {\scriptsize $\pm$ 0.032}} & \textbf{0.223 {\scriptsize $\pm$ 0.205}} & \textbf{0.142 {\scriptsize $\pm$ 0.037}} & \textbf{0.289 {\scriptsize $\pm$ 0.143}} \\
FlowDAS     & 0.233 {\scriptsize $\pm$ 0.075} & 0.331 {\scriptsize $\pm$ 0.229} & 0.283 {\scriptsize $\pm$ 0.154} & 0.411 {\scriptsize $\pm$ 0.163} & \uline{0.330 {\scriptsize $\pm$ 0.165}} & \uline{0.511 {\scriptsize $\pm$ 0.181}} \\
\bottomrule
\end{tabular}
\vspace{0.2in}
\caption{Lorenz-96 results on the quadratic capped observation operator at observation noise $\sigma_y = 0.2$ across state dimensions. Best results in \textbf{bold}; second best underlined.}
\label{tab:lorenz96_quad_capped_s02}
\vskip -0.1in
\end{table*}

\section{Full High-dimensional Lorenz-96 Results}
\label{app:localized_tables}
Complete results are in Table~\ref{tab:lorenz96_highdim_arctan_s02} and Table~\ref{tab:lorenz96_highdim_quartic_capped_s02}.
\begin{table*}[!t]
\vskip 0.1in
\centering
\small
\setlength{\tabcolsep}{5pt}
\begin{tabular}{@{}lcccc@{}}
\toprule
& \multicolumn{4}{c}{State Dimension $d_x$} \\
\cmidrule(l){2-5}
Method & 50 & 100 & 500 & 1000 \\
\midrule
\multicolumn{5}{@{}l}{\textit{RMSE} $\downarrow$} \\
\midrule
L-FPPF  & \textbf{0.287 $\pm$ 0.012} & \textbf{0.300 $\pm$ 0.014} & \textbf{0.293 $\pm$ 0.007} & \textbf{0.299 $\pm$ 0.004} \\
LETKF   & \underline{0.316 $\pm$ 0.022} & \underline{0.319 $\pm$ 0.015} & \underline{0.319 $\pm$ 0.007} & \underline{0.318 $\pm$ 0.004} \\
LBPF    & 0.366 $\pm$ 0.016 & 0.370 $\pm$ 0.011 & 0.372 $\pm$ 0.005 & 0.372 $\pm$ 0.004 \\
APF     & 0.669 $\pm$ 0.087 & 1.082 $\pm$ 0.183 & 2.386 $\pm$ 0.135 & 2.732 $\pm$ 0.089 \\
BPF     & 0.743 $\pm$ 0.122 & 1.232 $\pm$ 0.224 & 2.421 $\pm$ 0.146 & 2.776 $\pm$ 0.090 \\
EnKF    & 0.706 $\pm$ 0.133 & 1.187 $\pm$ 0.220 & 2.278 $\pm$ 0.162 & 2.522 $\pm$ 0.109 \\
\midrule
\multicolumn{5}{@{}l}{\textit{CRPS} $\downarrow$} \\
\midrule
L-FPPF  & \textbf{0.156 $\pm$ 0.006} & \textbf{0.162 $\pm$ 0.006} & \textbf{0.158 $\pm$ 0.003} & \textbf{0.161 $\pm$ 0.002} \\
LETKF   & \underline{0.177 $\pm$ 0.009} & \underline{0.178 $\pm$ 0.006} & \underline{0.177 $\pm$ 0.003} & \underline{0.177 $\pm$ 0.002} \\
LBPF    & 0.212 $\pm$ 0.007 & 0.213 $\pm$ 0.004 & 0.213 $\pm$ 0.002 & 0.213 $\pm$ 0.002 \\
APF     & 0.377 $\pm$ 0.053 & 0.707 $\pm$ 0.120 & 1.727 $\pm$ 0.103 & 2.021 $\pm$ 0.072 \\
BPF     & 0.439 $\pm$ 0.078 & 0.827 $\pm$ 0.150 & 1.757 $\pm$ 0.110 & 2.058 $\pm$ 0.070 \\
EnKF    & 0.411 $\pm$ 0.082 & 0.755 $\pm$ 0.144 & 1.599 $\pm$ 0.116 & 1.803 $\pm$ 0.084 \\
\bottomrule
\end{tabular}
\vspace{0.2in}
\caption{High-dimensional Lorenz-96 results with local filtering methods on the arctangent observation operator $o_{t,j}=\arctan(x_{t,j})+\epsilon$ at observation noise $\sigma_y = 0.2$. Best results in \textbf{bold}; second best underlined.}
\label{tab:lorenz96_highdim_arctan_s02}
\vskip -0.1in
\end{table*}

\begin{table*}[!t]
\vskip 0.1in
\centering
\small
\setlength{\tabcolsep}{5pt}
\begin{tabular}{@{}lcccc@{}}
\toprule
& \multicolumn{4}{c}{State Dimension $d_x$} \\
\cmidrule(l){2-5}
Method & 50 & 100 & 500 & 1000 \\
\midrule
\multicolumn{5}{@{}l}{\textit{RMSE} $\downarrow$} \\
\midrule
L-FPPF  & \textbf{0.670 $\pm$ 0.339} & \textbf{0.669 $\pm$ 0.245} & \textbf{0.698 $\pm$ 0.125} & \textbf{0.752 $\pm$ 0.075} \\
LETKF   & 1.935 $\pm$ 0.785 & 2.017 $\pm$ 0.640 & 2.152 $\pm$ 0.249 & 2.144 $\pm$ 0.175 \\
LBPF    & \underline{0.684 $\pm$ 0.226} & \underline{0.698 $\pm$ 0.254} & \underline{0.785 $\pm$ 0.132} & \underline{0.812 $\pm$ 0.085} \\
APF     & 1.908 $\pm$ 0.540 & 2.457 $\pm$ 0.356 & 3.087 $\pm$ 0.124 & 3.236 $\pm$ 0.087 \\
BPF     & 1.942 $\pm$ 0.552 & 2.534 $\pm$ 0.359 & 3.081 $\pm$ 0.124 & 3.245 $\pm$ 0.090 \\
EnKF    & 4.317 $\pm$ 0.538 & 4.623 $\pm$ 0.360 & 3.628 $\pm$ 0.222 & 3.318 $\pm$ 0.139 \\
\midrule
\multicolumn{5}{@{}l}{\textit{CRPS} $\downarrow$} \\
\midrule
L-FPPF  & \textbf{0.358 $\pm$ 0.158} & \textbf{0.338 $\pm$ 0.090} & \textbf{0.337 $\pm$ 0.043} & \textbf{0.349 $\pm$ 0.024} \\
LETKF   & 0.954 $\pm$ 0.439 & 0.951 $\pm$ 0.346 & 0.979 $\pm$ 0.159 & 0.970 $\pm$ 0.115 \\
LBPF    & \underline{0.366 $\pm$ 0.091} & \underline{0.366 $\pm$ 0.106} & \underline{0.375 $\pm$ 0.041} & \underline{0.383 $\pm$ 0.027} \\
APF     & 1.168 $\pm$ 0.369 & 1.633 $\pm$ 0.277 & 2.267 $\pm$ 0.106 & 2.421 $\pm$ 0.075 \\
BPF     & 1.213 $\pm$ 0.396 & 1.708 $\pm$ 0.278 & 2.268 $\pm$ 0.100 & 2.429 $\pm$ 0.072 \\
EnKF    & 3.065 $\pm$ 0.500 & 3.392 $\pm$ 0.337 & 2.629 $\pm$ 0.190 & 2.410 $\pm$ 0.117 \\
\bottomrule
\end{tabular}
\vspace{0.2in}
\caption{High-dimensional Lorenz-96 results with local filtering methods on the observation operator $\min(\bx_{t}^4,10)$ at observation noise $\sigma_y = 0.2$. Best results in \textbf{bold}; second best underlined.}
\label{tab:lorenz96_highdim_quartic_capped_s02}
\vskip -0.1in
\end{table*}

\clearpage

\end{document}